\def\CinemaTrajArxivVersion{}
\ifdefined\CinemaTrajArxivVersion
  \documentclass[sigconf,screen,nonacm]{acmart}
\else
  \documentclass[sigconf]{acmart}
\fi
\AtBeginDocument{%
  }

\ifdefined\CinemaTrajArxivVersion
  \setcopyright{none}
\else
  \copyrightyear{2026}
  \acmYear{2026}
  \setcopyright{cc}
  \setcctype{by}
  \acmConference[MM '26]{Proceedings of the 34th ACM International Conference on Multimedia}{November 10--14, 2026}{Rio de Janeiro, Brazil.}
  \acmBooktitle{Proceedings of the 34th ACM International Conference on Multimedia (MM '26), November 10--14, 2026, Rio de Janeiro, Brazil}
  \acmISBN{979-8-4007-2213-4/2026/11}
  \acmDOI{10.1145/3767308.3835366}
\fi

\ifdefined\CinemaTrajArxivVersion\else
  \acmSubmissionID{mfp2581}
\fi

\usepackage{multirow}
\usepackage{enumitem}
\usepackage{balance}
\newcommand{\degree}{\ensuremath{^\circ}}

\begin{document}

\newcommand{\model}{CinemaTraj}

\title{\model{}: Composing Atomic Camera Trajectories for 3D Scenes with LLM Agents}

\author{Qianru Li}
\authornote{Both authors contributed equally to this research.}
\orcid{0009-0002-5173-4306}
\affiliation{%
  \institution{Technical University of Munich}
  \city{Munich}
  \country{Germany}
}
\affiliation{%
  \institution{Huawei Dresden Research Center}
  \city{Munich}
  \country{Germany}
}
\email{qianru.li@tum.de}

\author{Xuyang Chen}
\authornotemark[1]
\orcid{0009-0004-4154-3463}
\affiliation{%
  \institution{Technical University of Munich}
  \city{Munich}
  \country{Germany}
}
\affiliation{%
  \institution{Huawei Dresden Research Center}
  \city{Munich}
  \country{Germany}
}
\email{xuyang.chen@tum.de}

\author{Erkin Türköz}
\orcid{0009-0005-1445-6839}
\affiliation{%
  \institution{Huawei Dresden Research Center}
  \city{Munich}
  \country{Germany}
}
\email{erkin.turkoz@huawei.com}

\author{Lu Liu}
\orcid{0009-0006-3173-1377}
\affiliation{%
  \institution{Huawei Dresden Research Center}
  \city{Munich}
  \country{Germany}
}
\email{luliu1@huawei.com}

\author{Xuqin Wang}
\orcid{0009-0008-4992-5342}
\affiliation{%
  \institution{Technical University of Munich}
  \city{Munich}
  \country{Germany}
}
\affiliation{%
  \institution{Huawei Dresden Research Center}
  \city{Munich}
  \country{Germany}
}
\email{xuqin.wang@tum.de}

\author{Liqiu Meng}
\orcid{0000-0001-8787-3418}
\affiliation{%
  \institution{Technical University of Munich}
  \city{Munich}
  \country{Germany}
}
\email{liqiu.meng@tum.de}

\author{Tao Wu}
\orcid{0009-0004-1519-6138}
\affiliation{%
  \institution{Huawei Dresden Research Center}
  \city{Munich}
  \country{Germany}
}
\email{taowu1@huawei.com}

\author{Yanfeng Zhang}
\authornote{Corresponding author.}
\orcid{0000-0003-1314-695X}
\affiliation{%
  \institution{Huawei Dresden Research Center}
  \city{Munich}
  \country{Germany}
}
\email{zhangyanfeng8@huawei.com}

\renewcommand{\shortauthors}{Qianru Li et al.}

\begin{abstract}
  Automatically generating cinematically expressive camera trajectories through 3D scenes from natural language descriptions is a challenging task of high practical value, with applications ranging from real-estate advertising to virtual tour creation. Existing methods either lack true 3D spatial awareness by relying on 2D image priors, or treat trajectory generation as a geometric path planning problem divorced from cinematographic semantics. We present \model{}, a framework that reframes camera trajectory planning as a language-grounded spatial reasoning problem. Given a set of RGB-D images and a user prompt, \model{} equips an LLM agent with a structured 3D scene graph: the agent decomposes the prompt into a sequence of atomic cinematographic movements (dolly, orbit, crane, pan, tilt, zoom, arc). Each movement is instantiated via a novel parametric trajectory representation that is both cinematographically expressive and optimizable for collision avoidance. The scene graph acts as a structured spatial prior, grounding the agent’s reasoning in accurate geometric and semantic knowledge of the environment. \model{} further generates synchronized voiceover and subtitles aligned with camera motion, producing narrated cinematic video outputs. We evaluate \model{} on real-world ScanNet++ environments, and show that it produces prompt-faithful, collision-free trajectories with high cinematographic quality, outperforming existing approaches on prompt alignment, trajectory quality, and safety metrics. Project webpage: \url{https://cinematraj.github.io/}.
\end{abstract}



\maketitle

\begin{figure*}[t]
  \centering
  \includegraphics[width=0.90\textwidth]{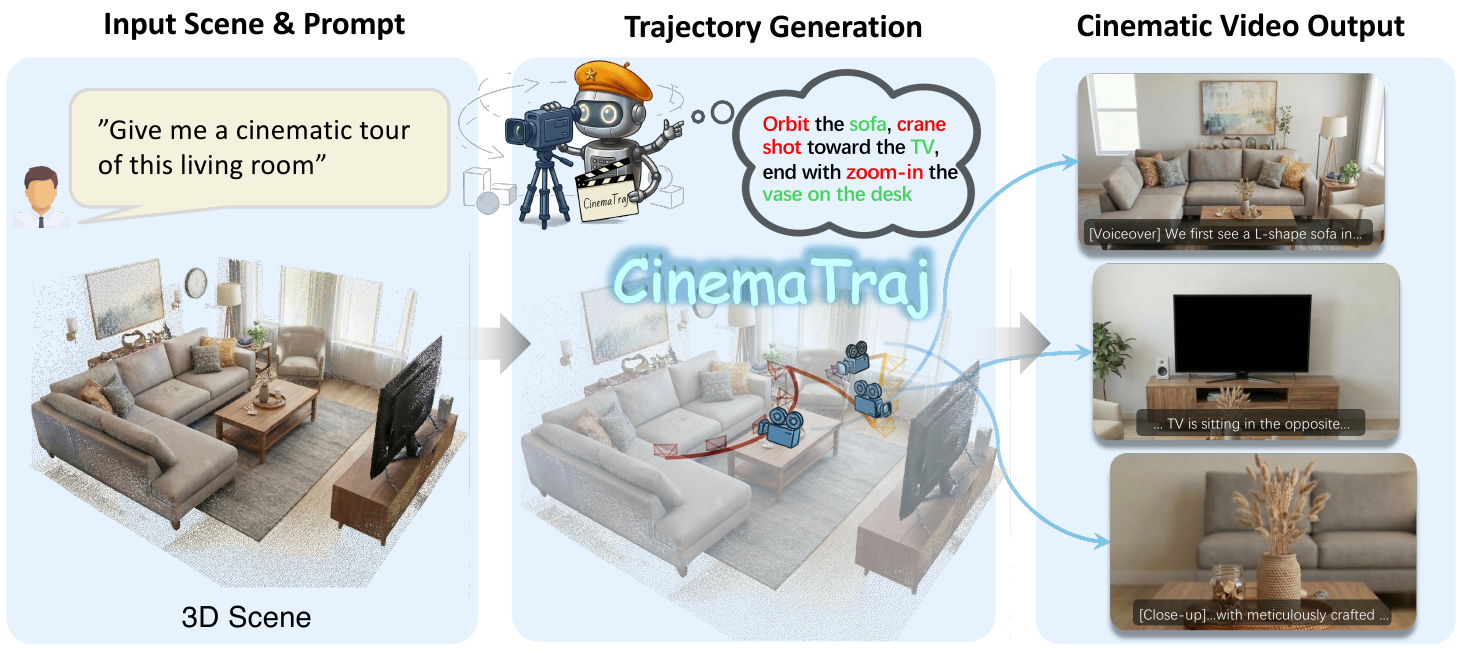}
    \caption{Given a 3D scene and a natural language prompt (\emph{left}), \model{} decomposes the request into a sequence of cinematographic movements --- orbit, crane, zoom-in --- grounded in a 3D scene graph, and plans a collision-free camera trajectory through the scene (\emph{middle}). The resulting trajectory is rendered into a cinematic video with synchronized voiceover and subtitles (\emph{right}).}
  \Description{Three side-by-side panels. Left panel, Input Scene and Prompt: a textured 3D reconstruction of a living room with an L-shaped sofa, a wall-mounted television, and a coffee table, plus a user speech bubble reading Give me a cinematic tour of this living room. Middle panel, Trajectory Generation: a robot director mascot converts the request into the plan orbit the sofa, crane shot toward the TV, end with zoom-in on the vase, and a smooth camera path with camera icons is drawn through the reconstructed room. Right panel, Cinematic Video Output: three rendered video frames, each with a subtitle bar carrying the synchronized voiceover text, linked by arrows to successive points along the trajectory.}
  \label{fig:teaser}
\end{figure*}

\section{Introduction}
Camera trajectory design targets generating continuous camera paths through 3D environments that are both cinematographically expressive and physically plausible.
In professional cinematography, each fundamental camera movement — dolly, orbit, crane, pan, tilt, zoom — carries specific visual semantics that shapes narrative mood and viewer perception~\cite{brown2022cinematography, mercado2022filmmakers}.
Designing such trajectories manually for architectural walkthroughs, product showcases, or establishing shots demands expert knowledge of when to push in for intimacy, orbit for spatial understanding, or pull back to establish context~\cite{brown2022cinematography}.
Automating this process from natural language instructions would unlock a range of practical applications, from real-estate advertisement generation to virtual tour creation and short-form video production.

Recent language-guided methods attempt to generate camera trajectories from text prompts, yet they operate without true 3D spatial awareness.
LLM-based approaches~\cite{chatcam} translate user intent into camera parameters using 2D image priors as placement references, but are not grounded in the scene's 3D geometry — they cannot perceive solid surfaces, leading to trajectories that frequently intersect scene structures.
Similarly, feedforward network-based methods~\cite{cinematographicdiffusion,e.t.,gendop} produce trajectories broadly aligned with prompts but do not account for the underlying 3D scene geometry, resulting in collisions with objects and failure to achieve object-level framing.

To address collision avoidance, recent 3D-aware methods~\cite{splatraj,gaussnav} leverage neural scene representations for path planning within 3D structures.
However, these approaches treat trajectory generation as a purely geometric optimization problem, divorced from cinematographic semantics — they can navigate around obstacles but cannot produce shots that carry professional visual meaning, such as a dolly-in for emphasis or an orbit for spatial revelation.
A core gap remains: no existing method jointly reasons about 3D scene structure and cinematographic intent from natural language descriptions.

We present \model{}, a framework that reframes camera trajectory planning in 3D scenes as a \emph{language-guided spatial reasoning} problem.
Our key insight is that by equipping an LLM agent with a structured 3D scene graph — encoding object identities, bounding boxes, room containment, and spatial relations — the agent can reason jointly about scene semantics and cinematographic intent.
Specifically, \model{} decomposes a user prompt into a sequence of atomic cinematographic movements, each realized through a novel parametric trajectory representation grounded in professional camera idioms.
This parametric design ensures that generated trajectories preserve the characteristic motion profiles of standard cinematic shots (e.g., smooth orbital arcs, gradual dolly approaches), unlike unconstrained pose-level generation that sacrifices cinematographic structure for geometric flexibility.

However, parametric trajectories initialized from scene-graph anchors may still collide with cluttered geometry or lose sight of target objects.
To address this, we introduce a differentiable collision and occlusion-free optimizer built on a signed distance field (SDF) constructed from the scene's bounding-box mesh.
The SDF provides smooth, geometry-faithful gradients that refine trajectory parameters while preserving shot style — a key advantage over prior methods~\cite{gaussnav, splatnav, foci} that rely on noisy 3DGS density fields for collision proxies.
Moreover, \model{} generates synchronized voiceover and subtitles aligned with camera motion via a VLM, enhancing the vividness and naturalness of the output.

Our technical contributions are summarized as follows:
\begin{itemize}[leftmargin=*, nosep]
\item We formalize the task of language-guided cinematic video generation from 3D scenes and introduce an end-to-end framework to address it.
\item We exploit 3D scene graphs as structured priors for spatial reasoning, providing the LLM agent with accurate geometric and semantic knowledge for trajectory planning, combined with a parametric trajectory representation grounded in cinematographic primitives, enabling precise prompt-aligned trajectory generation.
\item We propose an occlusion and collision-free shots optimization method specified for cinematographic shots creation. 
\item Extensive experiments demonstrate that our method generates trajectories that are collision-free, cinematographically grounded, and faithful to the user's intent, outperforming existing approaches across all evaluation axes.
\end{itemize}

\section{Related Work}

\begin{figure*}[!ht]
  \centering
  \includegraphics[width=\linewidth]{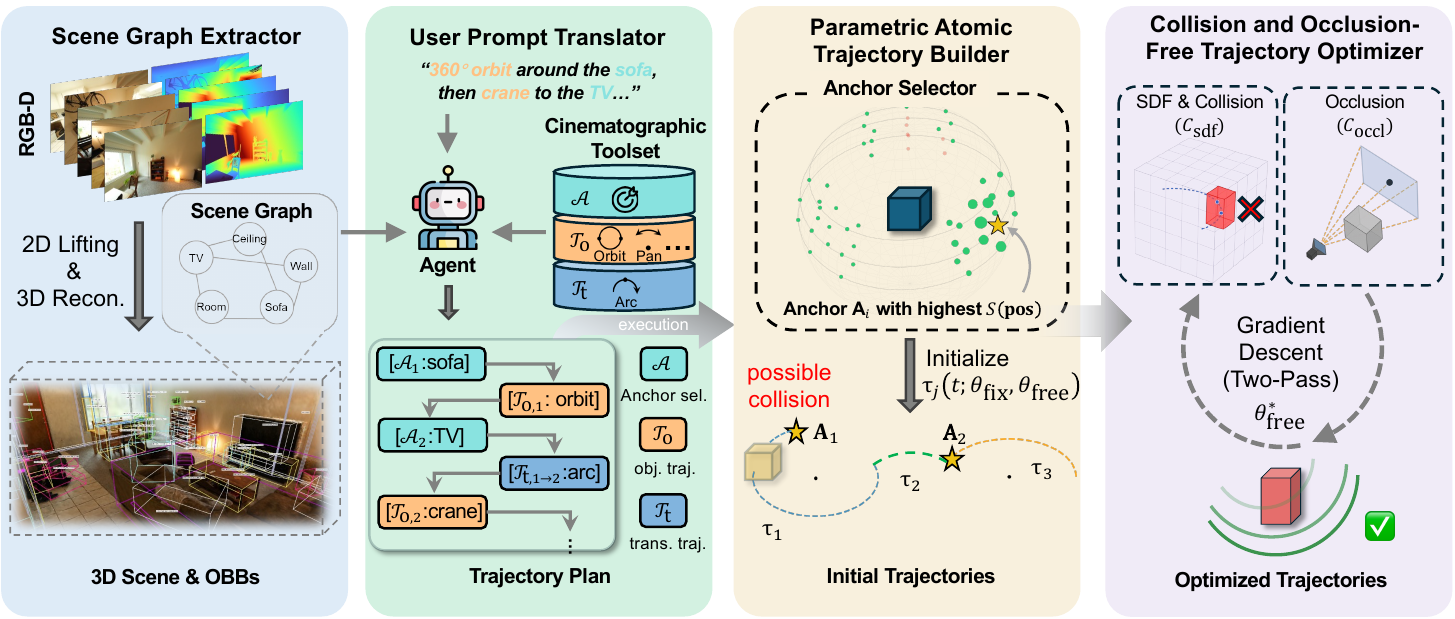}
     \caption{\textbf{Overview of \model{}.} The \emph{Scene Graph Extractor} takes RGB-D images as input and builds a hierarchical 3D scene graph with oriented bounding boxes (OBBs). Given a user prompt and the extracted scene graph, the \emph{User Prompt Translator} equips an LLM agent with a three-layer cinematographic toolset --- anchor selector~($\mathcal{A}$), object-level trajectories~($\mathcal{T}_\text{o}$), and transitional trajectories~($\mathcal{T}_\text{t}$) --- to decompose the prompt into a structured trajectory plan. This plan is then executed by the \emph{Parametric Atomic Trajectory Builder}, which selects optimal viewpoint anchors via scoring ($S(\mathbf{pos})$) and initializes each atomic command as a parametric trajectory $\tau_j(t;\,\boldsymbol{\theta}_\text{fix},\,\boldsymbol{\theta}_\text{free})$. Finally, the \emph{Collision and Occlusion-Free Trajectory Optimizer} refines $\boldsymbol{\theta}_\text{free}$ via two-pass gradient descent on SDF-based collision ($\mathcal{C}_\text{sdf}$) and occlusion ($\mathcal{C}_\text{occl}$) costs, producing optimized collision and occlusion-free trajectories.}
  \Description{Block diagram with four stages arranged left to right. Stage one, Scene Graph Extractor, converts a stack of RGB-D images via 2D lifting and 3D reconstruction into a 3D scene with oriented bounding boxes and a hierarchical scene graph whose nodes include room, ceiling, wall, TV, and sofa. Stage two, User Prompt Translator, feeds the user prompt to an LLM agent equipped with a three-layer cinematographic toolset of anchor selector, object-level trajectories, and transitional trajectories; the agent emits a structured trajectory plan that alternates anchor selections with atomic movements such as orbit, arc, and crane. Stage three, Parametric Atomic Trajectory Builder, scores candidate anchor positions around each target object, keeps the anchor with the highest score, and initializes every atomic command as a parametric trajectory; the initial trajectories may still pass through scene geometry. Stage four, Collision and Occlusion-Free Trajectory Optimizer, refines the free parameters with two-pass gradient descent on SDF-based collision and occlusion costs, producing final trajectories that clear all obstacles, indicated by a green check mark.}
  \label{fig:pipeline}
\end{figure*}

\subsection{Pose Optimization in 3D Scenes}
Camera trajectory generation in known 3D environments has traditionally been formulated as geometric optimization — solving constraint satisfaction over framing, visibility, and occlusion avoidance~\cite{christie2008camera, galvane2015camera, virtualcine, comprehensivesurvey, oskam2009visibility, bourne2005constraint}, or jointly optimizing aerial trajectory and orientation under obstacle constraints~\cite{nageli2017real}.
More recently, neural scene representations enable differentiable collision reasoning: Splat-Nav~\cite{splatnav} constructs safe corridors through 3DGS~\cite{3dgs} maps, FOCI~\cite{foci} formulates orientation-aware collision via Gaussian overlap integrals, and SplaTraj~\cite{splatraj} optimizes trajectories with differentiable framing and occlusion costs.
While these methods produce geometrically valid paths, they cannot interpret high-level cinematographic intent from natural language.

\subsection{Text-Conditioned Trajectory Generation}
Feedforward approaches~\cite{director3d, e.t., gendop, courant2025pulpmotion} train diffusion or autoregressive models to produce camera paths from text. To handle complex instructions, LLM-based planners decompose queries into sub-tasks: ChatCam~\cite{chatcam} localizes anchors via 2D image--text similarity, ACDC~\cite{acdc} extends this to aerial cinematography, WorldCraft~\cite{worldcraft} uses bounding-box placement in procedural worlds, FilmAgent~\cite{filmagent} selects from pre-authored camera setups, and LAMP~\cite{lamp} emits structured motion programs.
However, none are truly grounded in 3D geometry: feedforward models may penetrate walls, while LLM planners rely on 2D images~\cite{chatcam,acdc}, simplified abstractions~\cite{worldcraft}, or fixed stages~\cite{filmagent,lamp}.
In contrast, our method equips an LLM agent with a structured 3D scene graph, enabling joint reasoning over scene semantics, spatial layout, and cinematographic intent.

\subsection{3D Scene Graphs}
Scene graphs encode environments as graph-based abstractions with objects as nodes and spatial relationships as edges. Extended from 2D~\cite{johnson2015image, krishna2017visual} to 3D by Armeni~et~al.~\cite{armeni20193d}, subsequent work enables incremental construction from sensor streams~\cite{wu2021scenegraphfusion, hughes2022hydra, rosinol2021kimera}. Foundation-model-based methods achieve open-vocabulary generalization: ConceptGraphs~\cite{conceptgraphs} fuses SAM and CLIP into persistent 3D object graphs, and HOV-SG~\cite{hovsg} builds hierarchical open-vocabulary scene graphs for language-grounded navigation. Scene graphs also serve as planning interfaces for embodied agents~\cite{sayplan} and as control inputs for scene generation~\cite{graphdreamer, instructscene}.
We build on HOV-SG but apply 3D scene graphs to a previously unexplored setting: grounded spatial context for LLM-based cinematographic trajectory planning.

\section{Method}
\label{sec:method}

Given scene images (RGB-D) and a natural language prompt describing desired camera behavior, our system generates a collision-free, cinematographically plausible camera trajectory. The pipeline consists of four modules: (1)~a \textbf{Scene Graph Extractor} that builds an instance-level semantic graph of the scene, (2)~a \textbf{User Prompt Translator} that decomposes the user's intent into atomic camera movements via an LLM, (3)~a \textbf{Parametric Atomic Trajectory Builder} that instantiates these movements into a continuous trajectory, and (4)~a \textbf{Collision-Free Trajectory Optimizer} that refines trajectory parameters to avoid collisions and occlusions. An overview is shown in Figure~\ref{fig:pipeline}.

\subsection{Scene Graph Extractor}
\label{sec:scene_graph}

Standard 3D scene graphs~\cite{armeni20193d, hovsg} encode rich inter-object relations, yet for camera planning only two structural questions matter: \emph{which room contains the object}, and \emph{how is it attached to the room structure} (e.g., wall-mounted or ceiling-hung). These relations directly govern movement feasibility — a wall-adjacent object forbids full orbits; a ceiling object forbids crane shots. We therefore build a \emph{cinematography-oriented} graph retaining only room--object containment and object--structure edges.

\subsubsection{Graph Construction.}
Following HOV-SG~\cite{hovsg}, we build a hierarchical graph $\mathcal{G} = (\mathcal{N}, \mathcal{E})$ from RGB-D frames. Nodes $\mathcal{N} = \mathcal{N}_R \cup \mathcal{N}_O \cup \mathcal{N}_S$ (room, object, structure) each carry a semantic label~$\ell_i$ and an oriented bounding box (OBB) $\mathcal{B}_i = (\mathbf{c}_i, \mathbf{R}_i, \mathbf{e}_i)$. Per-frame 2D instance masks are lifted to 3D and merged into $M$ object nodes $\mathcal{N}_O = \{(\ell_i, \mathcal{B}_i)\}_{i=1}^{M}$. The edge set $\mathcal{E} = \mathcal{E}_{RO} \cup \mathcal{E}_{OS}$ encodes two relation types:
\begin{itemize}[leftmargin=*, nosep]
    \item \textbf{Room--object containment} $\mathcal{E}_{RO}$: each object is assigned to the room whose bounding box contains its center, establishing which spatial region the camera must navigate to reach the target.
    \item \textbf{Object--structure relation} $\mathcal{E}_{OS}$: each object is labeled $n_{S,i} \in \{\texttt{wall-adjacent},\, \texttt{ceiling-attached}\}$ based on its OBB proximity to wall surfaces or the ceiling plane. These labels directly constrain which camera movements are feasible around the object.
\end{itemize}

\subsubsection{Watertight Room Geometry.}
Adjacent structural bounding boxes (walls, floor, ceiling) are extended and snapped to form a \emph{watertight} enclosure, ensuring the signed distance field (\S\ref{sec:optimizer}) has no leakage through room boundaries.

\subsection{User Prompt Translator}
\label{sec:prompt_translator}

To bridge the gap between free-form text and executable camera commands, we provide an LLM with the scene graph, a predefined cinematographic toolset, and a dialogue protocol, enabling it to translate prompts into a sequence of atomic trajectory commands.

\subsubsection{Cinematographic Toolset.}
We define the object viewpoint selection tool and two categories of camera movements here; see \S\ref{sec:parametric_builder} for detailed explanation:

\begin{itemize}[leftmargin=*, nosep]
    \item \textbf{Anchor Selector} $\mathcal{A}$: selects a collision/occlusion-free initial viewpoint for each target object via a score-based mechanism (details in \S\ref{sec:parametric_builder}).
    \item \textbf{Object-level trajectories} $\mathcal{T}_\text{o}$: movements that examine a specific object, including orbital shots (\texttt{orbit full, orbit half, orbit quarter}), panning and tilting (\texttt{pan left, pan right}, \texttt{tilt up, tilt down}), dolly (\texttt{move in, move out}), optical zoom (\texttt{zoom in\_out, zoom out\_in}), vertical movements (\texttt{crane}), and a stationary \texttt{static} hold.
    \item \textbf{Transitional trajectories} $\mathcal{T}_\text{t}$: movements transporting the camera between objects. We use a parameterized \texttt{arc} with curvature $\alpha \in [-89\degree, 89\degree]$, where $\alpha{=}0$ yields a straight line and larger values create sweeping curves.
\end{itemize}

The LLM also specifies \textbf{viewing preferences} per object: elevation $e \in \{\texttt{low}, \texttt{medium}, \texttt{high}, \texttt{overhead}\}$ and framing distance $d \in \{\texttt{close}, \texttt{medium}, \texttt{far}\}$, which guide anchor placement. The object--structure relation further constrains trajectory selection (e.g., wall objects forbid \texttt{orbit full}; ceiling objects forbid \texttt{crane}).

\subsubsection{Prompt to Atomic Commands.}
The LLM receives the scene graph $\mathcal{G} = (\mathcal{N}, \mathcal{E})$, the user request, and the toolset $\mathcal{A}$, $\mathcal{T}_\text{o} \cup \mathcal{T}_\text{t}$. For a sequence of $N$ objects $(o_1, o_2, \ldots, o_N)$, the plan produces $3N - 1$ atomic steps:
\begin{equation}
\begin{split}
    \underbrace{\mathcal{A}_1}_{\text{start}} \;\rightarrow\;
    \underbrace{\mathcal{T}_{\text{o},1} \rightarrow \mathcal{A}_2 \rightarrow \mathcal{T}_{\text{t},{1 \rightarrow 2}} \rightarrow \mathcal{T}_{\text{o},2} \rightarrow \cdots}_{\text{loop}} \\
    \underbrace{\cdots \rightarrow \mathcal{A}_N \rightarrow \mathcal{T}_{\text{t},{(N{-}1) \rightarrow N}}}_{\text{loop (cont.)}} \;\rightarrow\;
    \underbrace{\mathcal{T}_{\text{o},N}}_{\text{end}}\text{,}
\end{split}
\end{equation}
where $\mathcal{A}_i$ is an anchor selector, $\mathcal{T}_{\text{o},i}$ is an object-level trajectory for $o_i$, and $\mathcal{T}_{\text{t},{i \rightarrow j}}$ is a transitional arc between $o_i$ and $o_j$. Each step is a structured command (tool name, target object ID, parameters), decoupling intent understanding from trajectory generation.

\subsection{Parametric Atomic Trajectory Builder}
\label{sec:parametric_builder}

The Trajectory Builder instantiates each atomic command into a continuous camera path using a parametric representation: each trajectory type is defined by a small set of interpretable parameters, preserving cinematographic motion profiles while enabling gradient-based collision optimization (\S\ref{sec:optimizer}).

\subsubsection{Anchor Determination.}
For each target object $o_i$, the Anchor Selector generates candidate viewpoints $\{\mathbf{A}_i^{(j)}\}_{j=1}^{N_i}$, each defined as $\mathbf{A}_i^{(j)} = (\mathbf{pos}_i^{(j)}, \mathbf{c}_i, S_i^{(j)})$ (position, look-at point, quality score). Candidates are sampled via \textbf{face-normal-biased sampling} on accessible OBB faces at the LLM-specified elevation and distance. After SDF-based collision filtering (\S\ref{sec:optimizer}), the best anchor is selected by:
\begin{equation}
\begin{split}
    S(\mathbf{pos}) = & w_\text{vis}\, S_\text{vis} + w_\text{dist}\, S_\text{dist} + w_\text{elev}\, S_\text{elev} \\
    & + w_\text{open}\, S_\text{open} + w_\text{perp}\, S_\text{perp}\text{,}
    \label{eq:anchor_score}
\end{split}
\end{equation}
where $S_\text{vis}$ measures bounding-box corner visibility via ray casting, $S_\text{dist}$ penalizes deviation from preferred distance, $S_\text{elev}$ scores elevation angle, and $S_\text{open}$ favors positions near the room center. For wall-adjacent objects, $S_\text{perp}$ additionally rewards viewpoints facing from the room interior; otherwise $w_\text{perp}{=}0$.

\subsubsection{Parametric Representation.}
 Each atomic trajectory $\tau$ has \textbf{fixed parameters} $\boldsymbol{\theta}_\text{fix}$ (from anchors/geometry) and \textbf{free parameters} $\boldsymbol{\theta}_\text{free}$ (optimizable):
\begin{equation}
    \tau(t; \boldsymbol{\theta}_\text{fix}, \boldsymbol{\theta}_\text{free}) = \big(\mathbf{p}(t),\; \mathbf{R}(t)\big) \in \mathbb{R}^3 \times SO(3)\text{,}
\end{equation}
All trajectories use ease-in-out interpolation $\hat{t} = \sigma(t)$ for smooth acceleration.

\paragraph{Object-level Trajectories.}
The camera orientation is always $\mathbf{R}(t) = \texttt{look\_at}(\mathbf{p}(t), \mathbf{c}_i)$, where $\mathbf{c}_i$ is the target object center, with position parameterized per shot type:

\begin{itemize}[leftmargin=*, nosep]
    \item \textbf{Orbit}: Circular arc at height $h$, radius $r$, sweeping angle $\phi_s \to \phi_e$. Free parameters: $\boldsymbol{\theta}_\text{free} = (r, h, \phi_s, \phi_e, \delta_\text{pitch})$ with $\delta_\text{pitch}$ as an optional pitch offset. The sweep angles are set to $360\degree$, $180\degree$, or $90\degree$ for full, half, and quarter orbits respectively.

    \item \textbf{Dolly}: Radial translation along approach direction $\hat{\mathbf{d}}$, interpolating between start radius $r_s$ and end radius $r_e$.

    \item \textbf{Crane}: Vertical arc from elevation $\epsilon_s$ to $\epsilon_e$ at fixed azimuth and radius $r$.

    \item \textbf{Pan/Tilt}: Stationary camera with yaw or pitch interpolation.

    \item \textbf{Optical Zoom}: Stationary camera with focal length $f(t) = f_0 \cdot [1 + (m_\text{peak} - 1)\sin(\pi \hat{t})]$; the sinusoidal envelope returns to $f_0$ at both endpoints.

    \item \textbf{Static}: Camera holds its anchor pose.
\end{itemize}

\paragraph{Transitional Trajectory.}
The \textbf{Arc} connects two anchors $(\mathbf{p}_s, \mathbf{Q}_s)$ and $(\mathbf{p}_e, \mathbf{Q}_e)$ with a single free parameter---the arc angle $\alpha$---controlling lateral displacement:
\begin{equation}
    \mathbf{p}(t) = (1 {-} \hat{t})\,\mathbf{p}_s + \hat{t}\,\mathbf{p}_e
    + \tfrac{L}{2}\tan(\alpha)\sin(\pi\,\hat{t})\;\hat{\mathbf{n}}\text{,}
\end{equation}
where $L = \|\mathbf{p}_e - \mathbf{p}_s\|$ and $\hat{\mathbf{n}}$ is perpendicular to the chord in the horizontal plane.

\subsubsection{Trajectory Combination.}
Atomic segments are concatenated sequentially; positional continuity is guaranteed by construction since each arc's endpoints match the surrounding anchors. When no visitation order is specified, a nearest-neighbor heuristic minimizes total camera travel.

\subsection{Collision and Occlusion-Free Trajectory Optimizer}
\label{sec:optimizer}
Initialized trajectories may still collide with furniture or lose sight of targets. We refine $\boldsymbol{\theta}_\text{free}$ via differentiable optimization on a signed distance field (SDF), which provides geometrically faithful gradients — unlike the 3DGS density fields used by prior work~\cite{gaussnav, splatnav, foci} that are noisy proxies for solid geometry.

\subsubsection{SDF Construction.}
We compute a discretized SDF $\Phi: \mathbb{R}^3 \rightarrow \mathbb{R}$ over a voxel grid from the OBB mesh, with differentiable trilinear interpolation enabling gradient flow to trajectory parameters.

\subsubsection{Cost Formulation.}
For each atomic trajectory $\tau_k$, we uniformly sample positions and minimize:
\begin{equation}
    \boldsymbol{\theta}_{\text{free},k}^* = \arg\min_{\boldsymbol{\theta}_{\text{free},k}} \; \mathcal{C}_\text{sdf} + \mathcal{C}_\text{reg} + \mathcal{C}_\text{occl} + \mathcal{C}_\text{bnd}\text{.}
    \label{eq:total_cost}
\end{equation}
The \textbf{collision cost} penalizes points penetrating or approaching scene geometry:
\begin{equation}
    \mathcal{C}_\text{sdf} = \frac{\lambda_\text{sdf}}{|\mathcal{P}|} \sum_{t_j \in \mathcal{P}} \big[\rho(g_j) + \rho(g_j)^2\big], \quad g_j = d_\text{safe} - \Phi(\mathbf{p}(t_j))\text{,}
\end{equation}
where $\rho(x) = \max(0, x)$. $\mathcal{C}_\text{reg}$ penalizes deviation from initial parameters (range-normalized), and $\mathcal{C}_\text{bnd}$ applies exponential barriers at scene boundaries.

\paragraph{Occlusion Cost.}
For each camera position $\mathbf{p}(t_j)$, we cast a ray toward the target center and march in $K$ steps, querying the SDF. Samples within the target's OBB are masked to avoid self-occlusion. Occlusion is accumulated via transmittance weighting:
\begin{equation}
    \mathcal{C}_\text{occl} = \sum_{m=1}^{K} T_m \cdot h_m, \quad T_m = \exp\!\Big(-\sum_{j=1}^{m-1} h_j\Big)\text{,}
    \label{eq:occlusion}
\end{equation}
where $h_m = \mathrm{softplus}_\beta(\delta_\text{occ} - \Phi(\mathbf{r}_j(s_m)))$ and the transmittance $T_m$ ensures the first occluder dominates the penalty. Both ray origins and SDF queries are differentiable, enabling end-to-end gradient flow.

\subsubsection{Two-Pass Optimization.}
\textbf{Pass~1} optimizes object-level segments with the full cost (including occlusion), operating on a per-type subset of $\boldsymbol{\theta}_\text{free}$ to preserve shot style. \textbf{Pass~2} rebuilds transitional arcs from updated endpoints and optimizes without occlusion cost; cross-room arcs are split into sub-arcs through door waypoints from the room connectivity graph. By only adjusting $\boldsymbol{\theta}_\text{free}$, the optimizer avoids obstacles while preserving shot-defining structure — an orbit remains an orbit even in cluttered scenes.

\subsection{Subtitle and Voiceover Generator}
\label{sec:generator}

We optionally generate subtitles and voiceover to narrate the camera walkthrough. The rendered video is segmented into short intervals; keyframes are extracted and fed to a vision-language model, which outputs (1)~\textbf{subtitles}---one caption per segment, and (2)~\textbf{voice-} \textbf{over blocks}---longer narration spans covering 2--4 segments. Each voiceover block is synthesized via TTS, temporally fitted to its window, and concatenated into a synchronized audio track.

\section{Experiments}
\label{sec:experiments}

\begin{figure*}[!ht]
  \centering
  \includegraphics[width=0.76\linewidth]{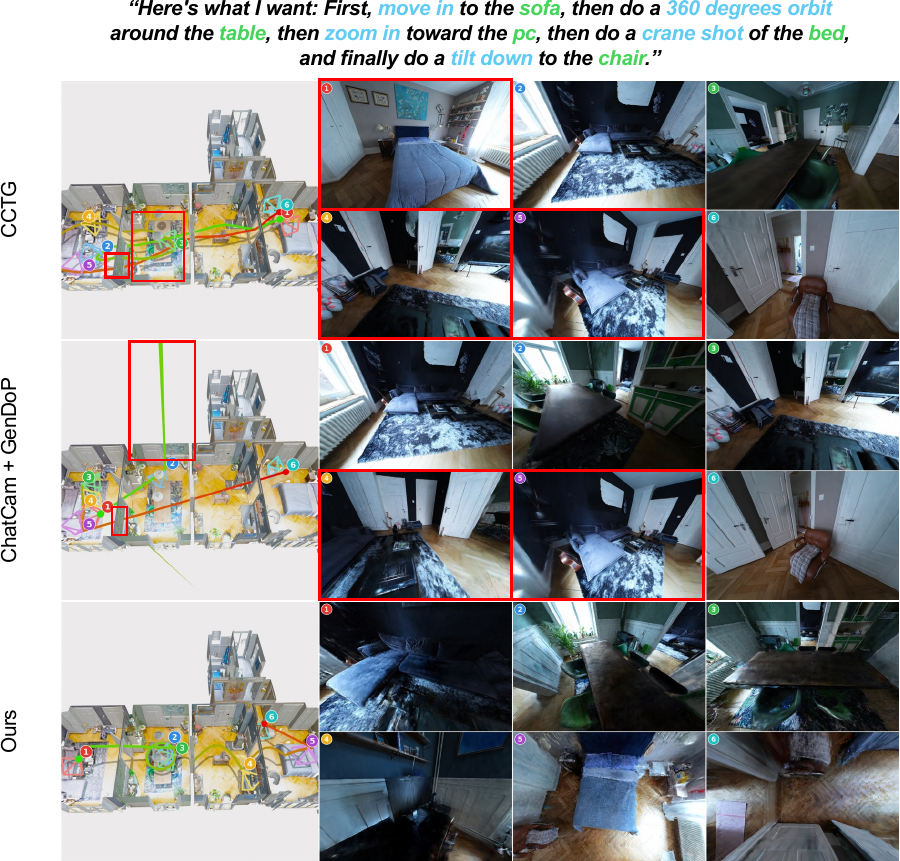}
  \caption{\textbf{Qualitative comparison} of trajectory visualizations and rendered keyframes on ScanNet++ scenes. Our method produces smooth, collision-free trajectories that faithfully follow the requested camera movements, while both baselines penetrate scene geometry and ChatCam + GenDoP exhibits erratic motions.}
  \Description{Three rows of results, labeled CCTG, ChatCam plus GenDoP, and Ours, for the same prompt asking to move in to the sofa, orbit 360 degrees around the table, zoom in toward the PC, take a crane shot of the bed, and tilt down to the chair. Each row pairs a top-down view of the reconstructed apartment with the generated camera path drawn on it, and six rendered keyframes taken along the path; red outlines mark failure frames. In the CCTG row the path cuts through walls and several red-outlined keyframes are dark or clipped inside geometry. In the ChatCam plus GenDoP row the path contains long straight jumps that leave the apartment footprint and multiple red-outlined keyframes show wall penetrations. In the Ours row the path stays inside free space, visits the prompted objects in order, and all six keyframes show clean, unobstructed views of the requested targets.}
  \label{fig:comparison}
\end{figure*}

\begin{figure*}[!ht]
  \centering
  \includegraphics[trim=0 -10 0 -10,width=0.76\linewidth]{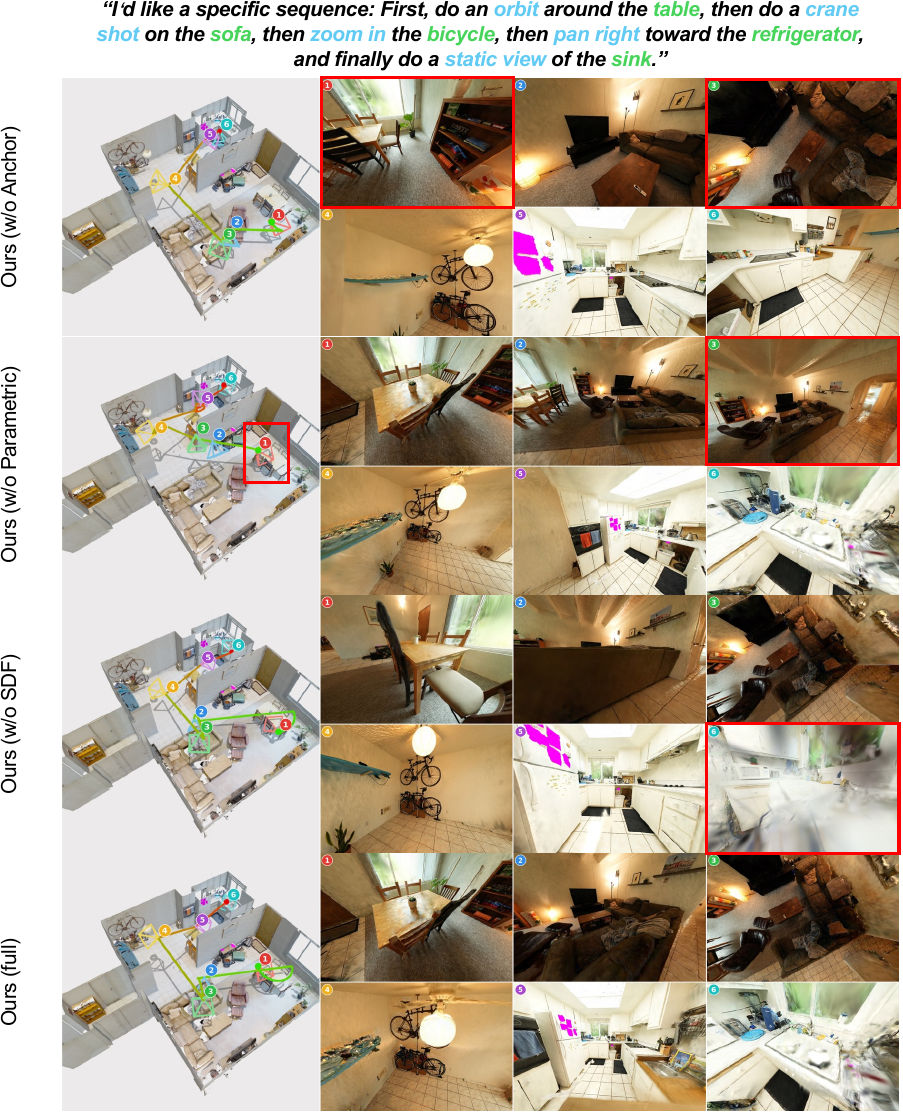}
  \caption{\textbf{Ablation visualization.} Removing the Anchor Selector leads to incorrect object targeting; removing parametric trajectories yields unstructured paths; removing SDF optimization causes geometry collisions. The full pipeline combines all three strengths.}
  \Description{Four rows compare ablated variants on one apartment scene for a prompt requesting an orbit around the table, a crane shot on the sofa, a zoom in on the bicycle, a pan right toward the refrigerator, and a static view of the sink. Each row pairs a top-down trajectory map with six rendered keyframes, and red outlines mark failures. Without the Anchor Selector, the camera targets wrong objects and several keyframes frame incorrect content. Without parametric trajectories, the path is jagged and unstructured and a keyframe misses its target. Without SDF optimization, the path clips into furniture and one keyframe is filled with gray collision geometry. The full pipeline row shows a smooth path visiting all five requested targets with clean keyframes throughout.}
  \label{fig:ablation}
\end{figure*}

\subsection{Experimental Setup}
\label{sec:exp_setup}

\subsubsection{Implementation Details.}
We use GPT-4.1~\cite{gpt4} as the LLM backbone for both the User Prompt Translator and the Subtitle and Voiceover Generator throughout all experiments. For anchor scoring in Equation~\ref{eq:anchor_score}, we set $w_\text{vis}=0.35$, $w_\text{dist}=0.20$, $w_\text{elev}=0.10$, $w_\text{open}=0.10$, and $w_\text{perp}=0.25$ for wall-adjacent objects; for free-standing objects, we set $w_\text{vis}=0.45$, $w_\text{dist}=0.25$, $w_\text{elev}=0.15$, $w_\text{open}=0.15$, and $w_\text{perp}=0$. The SDF voxel grid is computed at $256^3$ resolution, and the two-pass trajectory optimizer runs Adam for 1000 iterations per segment with an initial learning rate of 1.0.

\subsubsection{Dataset and Evaluation Protocol.}
We evaluate on 50 scenes from ScanNet++~\cite{yeshwanth2023scannet++}, a real-world indoor dataset with high-fidelity 3D reconstructions. For each scene we generate two prompts at each of three specificity levels, yielding six prompts per scene:
\begin{itemize}[leftmargin=*, nosep]
    \item \textbf{Fully-specified} prompts designate both target objects and exact camera movements (\emph{e.g.}, ``First, do a full 360° orbit around the plant, then tilt up from the display stand''), testing faithful execution of explicit cinematographic instructions.
    \item \textbf{Partially-specified} prompts name ${\geq}3$ objects but leave movement selection to the system (\emph{e.g.}, ``Highlight the dining plate, bread, plant, and vase in a smooth sequence''), testing automatic shot planning given known targets.
    \item \textbf{Open-ended} prompts give abstract cinematic requests with no object or movement specification (\emph{e.g.}, ``Create a cinematic tour of this interior''), testing end-to-end scene understanding and planning.
\end{itemize}
Prompts are generated via template-based sampling with a fixed seed for reproducibility. We evaluate on \textbf{fully-specified prompts} for the main comparison and ablation: because these prompts specify both target objects and exact camera movements, they provide unambiguous ground-truth intent against which all methods — including those without scene-graph access — can be fairly assessed. Partially-specified and open-ended results, which test autonomous planning capability, are reported in the supplementary material.

\subsubsection{Baselines.}
We compare against the two most relevant publicly available methods, representing complementary paradigms:
\begin{itemize}[leftmargin=*, nosep]
    \item \textbf{ChatCam~\cite{chatcam}\,+\,GenDoP~\cite{gendop}}: ChatCam's prompt translator and anchor selector, combined with GenDoP's text-conditioned diffusion model for 6-DoF trajectory generation. This baseline operates \emph{without} access to the 3D scene graph or geometry — the trajectory is synthesized from text and 2D images, representing the text-conditioned generation paradigm.
    \item \textbf{CCTG}~\cite{wu2025cinematographic}: A language-guided camera trajectory generation pipeline with collision awareness. This baseline \emph{has} access to the 3D scene's geometry and relies on keyframe interpolation for trajectory generation, but without explicit cinematographic constraints, representing the geometric optimization paradigm.
\end{itemize}
All methods receive identical prompts and, where applicable, identical scene renderings. Other recent methods~\cite{e.t.,director3d,courant2025pulpmotion} either do not operate in 3D scenes or lack publicly available code for fair reproduction.

\subsubsection{Metrics.}
We evaluate along five complementary axes:
\begin{itemize}[leftmargin=*, nosep]
    \item \textbf{Motion MSE}: For fully-specified prompts, proposed in~\cite{chatcam}, we compute the mean squared error between the generated trajectory and a hand-crafted ground-truth trajectory created in Blender. Both trajectories are uniformly resampled by arc length to a fixed number of points.
    \item \textbf{CLaTr Score}~\cite{e.t.}: Cosine similarity between trajectory and text embeddings in the shared CLaTr latent space ($\uparrow$).
    \item \textbf{Collision Rate}: Fraction of trajectory samples where the SDF value falls below zero ($\downarrow$).
    \item \textbf{Occlusion Rate}: Fraction of trajectory samples where the target object is blocked ($\downarrow$).
    \item \textbf{Object Coverage}: Fraction of planned anchor objects successfully visited ($\uparrow$).
\end{itemize}

\subsection{Quantitative Results}
\label{sec:quant}

\subsubsection{Comparison with Baselines.}
Table~\ref{tab:comparison&ablation} reports results on the ScanNet++ benchmark using fully-specified prompts. Our method achieves the best performance across all metrics except collision rate, where we rank second to the w/o Anchor ablation variant (0.056 vs.\ 0.023).
ChatCam+GenDoP, operating without scene geometry, produces the highest collision rate (0.209) and occlusion rate (0.580) among all methods — the diffusion model cannot reason about solid surfaces, leading to frequent geometry penetration and loss of target visibility.
CCTG first synthesizes a video and then recovers camera poses via Structure-from-Motion (SfM). While the SfM reconstruction yields a relatively low collision rate (0.035), the recovered trajectories are inherently noisy, resulting in the worst Motion MSE (9.143) across all methods despite moderate semantic alignment (CLaTr 24.031).

\subsubsection{Ablation Study.}

We ablate three core components (Table~\ref{tab:comparison&ablation}, lower section, Fig.~\ref{fig:ablation}):
\begin{itemize}[leftmargin=*, nosep]
    \item \textbf{Anchor Selector}: Replacing our scene-graph-based module with CLIP-based anchor selection preserves low collision rates (0.023) but substantially degrades motion fidelity (MSE 7.055 vs.\ 1.741) and coverage (0.882 vs.\ 1.000), as CLIP occasionally selects visually similar but spatially incorrect targets.
    \item \textbf{Parametric Trajectories}: Replacing our parametric representation with direct 6-DoF diffusion generation achieves the best occlusion rate (0.460), as unconstrained poses can freely orient toward targets. However, collision rate increases to 0.081 and motion fidelity degrades significantly (MSE 2.843 vs.\ 1.741), confirming that parametric templates trade pose-level flexibility for cinematographically coherent, physically plausible motion.
    \item \textbf{SDF-based Optimization}: Falling back to the 3DGS density field causes the collision rate to spike to 0.557 — the worst across all methods — confirming that raw Gaussian densities are a poor proxy for solid geometry. Motion fidelity remains competitive (MSE 1.841), indicating that this module primarily governs spatial safety rather than trajectory shape.
\end{itemize}
The full pipeline combines the strengths of all three components: the Anchor Selector for correct grounding, parametric trajectories for cinematographic structure, and SDF-based optimization for physical plausibility.

\begin{table}[t]
\centering
\caption{\textbf{Quantitative comparison and ablation} on ScanNet++. Best in \textbf{bold}, second best \underline{underlined}.}
\label{tab:comparison&ablation}
\setlength{\tabcolsep}{4pt}
\resizebox{\linewidth}{!}{%
\begin{tabular}{l ccccc}
\toprule
\multirow{2}{*}{\textbf{Method}} & \textbf{Motion} & \textbf{CLaTr} & \textbf{Collision} & \textbf{Occlusion} & \textbf{Coverage} \\
 & \textbf{MSE}\,$\downarrow$ & \textbf{Score}\,$\uparrow$ & \textbf{Rate}\,$\downarrow$ & \textbf{Rate}\,$\downarrow$ & \textbf{Rate}\,$\uparrow$ \\
\midrule
ChatCam + GenDoP~\cite{chatcam,gendop}   & 7.741             & 19.461             & 0.209             & 0.580             & 0.661             \\
CCTG~\cite{wu2025cinematographic}        & 9.143             & 24.031             & \underline{0.035} & 0.516             & 0.700             \\
\midrule
w/o Anchor Selector                  & 7.055             & \underline{28.538}             & \textbf{0.023}    & 0.579             & \underline{0.882} \\
w/o Parametric Traj.                     & 2.843             & 26.243    & 0.081             & \textbf{0.460}    & 1.000             \\
w/o SDF-based Opt.                       & \underline{1.841} & 25.900 & 0.557             & 0.566             & 1.000             \\
\textbf{\model{} (Ours)}                 & \textbf{1.741}    & \textbf{28.982}    & 0.056             & \underline{0.503} & \textbf{1.000}    \\
\bottomrule
\end{tabular}%
}
\end{table}

\begin{table}[t]
\centering
\caption{\textbf{User study}}
\label{tab:user_study}
\setlength{\tabcolsep}{3.5pt}
\resizebox{\linewidth}{!}{%
\begin{tabular}{l ccc}
\toprule
\textbf{Method}          & \textbf{Prompt Align.}\,$\uparrow$ & \textbf{Collision \& Occlusion Avoid.}\,$\uparrow$ & \textbf{Cine.\ Quality}\,$\uparrow$ \\
\midrule
CCTG                     & 2.38                               & 2.93                                               & 2.16                                \\
ChatCam + GenDoP         & 2.00                               & 1.97                                               & 2.08                                \\
\midrule
w/o Anchor Selector  & 3.03                               & 3.63                                               & 3.24                                \\
w/o Parametric Traj.     & 2.89                               & 3.40                                               & 2.14                                \\
w/o SDF-based Opt.       & 3.17                               & 2.01                                               & 2.60                                \\
\textbf{\model{} (Ours)} & \textbf{4.62}                      & \textbf{4.56}                                      & \textbf{4.36}                       \\
\bottomrule
\end{tabular}%
}
\end{table}

\subsubsection{User Study.}
We conduct a human evaluation in which 40 participants view trajectory-rendered videos from all methods (including ablation variants) across 4 representative scenes. Each video is rated on a 5-point Likert scale along three criteria: \emph{Prompt Alignment}, \emph{Collision \& Occlusion Avoidance}, and \emph{Cinematographic Quality}. Table~\ref{tab:user_study} reports the results. Our method is preferred across all three criteria, with particularly strong gains in cinematographic quality---reflecting the benefit of the parametric shot vocabulary---and collision avoidance.

\subsection{Qualitative Results}
\label{sec:qual}

\subsubsection{Visual Comparison.}
Fig.~\ref{fig:comparison} shows trajectory visualizations and rendered keyframes for all methods on selected scenes. Our trajectories exhibit smooth, cinematographically structured motion that closely follows the requested camera movements while maintaining safe clearance from scene geometry. In contrast, ChatCam+GenDoP produces trajectories that both collide with furniture and exhibit erratic motion, as its diffusion model lacks access to 3D scene structure. CCTG avoids most collisions but yields jagged, non-smooth trajectories that lack the fluidity characteristic of professional cinematographic camera work.

\subsubsection{Ablation Visualization.}
Fig.~\ref{fig:ablation} illustrates the effect of each ablated component. Without the Anchor Selector, the system occasionally targets incorrect objects; without parametric trajectories, the generated paths lack cinematographic structure; without SDF optimization, trajectories collide with scene geometry. The full pipeline combines all three strengths.

\subsubsection{Shot Diversity and Challenging Scenarios.}
Fig.~\ref{fig:comparison} shows examples of several shot types supported by our parametric representation, including orbits, crane shots, dolly moves, and pans/tilts. The parametric formulation preserves each shot's characteristic motion profile even after the collision-free optimization adjusts trajectory parameters. We also test on more demanding settings: densely furnished rooms where collision avoidance is frequently triggered, and multi-object sequences that require the planner to chain multiple shot types. Additional examples, including medium- and high-level prompt results, are provided in the Appendix.

\section{Conclusion and Limitation}
We presented \model{}, a framework that reframes camera trajectory planning in 3D scenes as a language-guided spatial reasoning problem. The key idea is to equip an LLM agent with a structured 3D scene graph, enabling joint reasoning over scene semantics and cinematographic intent. Combined with a parametric trajectory representation grounded in professional camera idioms and an SDF-based differentiable optimizer for collision and occlusion avoidance, \model{} produces prompt-faithful, collision-free trajectories with high cinematographic quality. Experiments on ScanNet++ demonstrate consistent improvements over existing approaches across motion fidelity, semantic alignment, collision avoidance, and object coverage, confirmed by a 40-participant user study in which our method is strongly preferred on all evaluation criteria.

However, limitations remain. Our framework currently focuses on one-shot trajectory generation in static 3D scenes. Extending to dynamic scenes with moving actors would require the planner to jointly reason about camera and character trajectories over time, maintaining proper framing as spatial configurations change. Our parametric trajectory library also covers only foundational movement types; subject-relative shots such as tracking, over-the-shoulder, and point-of-view compositions remain to be addressed. Finally, generating multi-shot sequences with cuts introduces the need for editing grammar---enforcing the 180-degree rule, shot--reverse-shot patterns, and eyeline continuity---bridging the gap from single-trajectory generation to full cinematic storytelling. We leave these directions for future work.

\bibliographystyle{ACM-Reference-Format}
\balance
\bibliography{base}

\ifdefined\CinemaTrajArxivVersion
  \clearpage
  \appendix
  \twocolumn[{%
    \centering
    {\Large\bfseries Supplementary Material\par}
    \vspace{0.75em}
  }]
  \addcontentsline{toc}{section}{Supplementary Material}
  \suppressfloats[t]
  \counterwithin{figure}{section}
  \counterwithin{table}{section}
  \appendix
\renewcommand{\thefigure}{\thesection.\arabic{figure}}
\renewcommand{\thetable}{\thesection.\arabic{table}}

\section{Video Results}
We provide supplementary video results to complement the static comparisons in the main paper. The videos are organized into three parts:
\begin{itemize}[nosep,leftmargin=1.5em]
  \item \textbf{Baseline comparison}: side-by-side rendered trajectories from \model{}, ChatCam+GenDoP~\cite{chatcam,gendop}, and CCTG~\cite{wu2025cinematographic} on the same ScanNet++~\cite{yeshwanth2023scannet++} scenes shown in Fig.~3. The videos reveal temporal artifacts (jitter, abrupt heading changes) that are difficult to assess from keyframes alone.
  \item \textbf{Ablation comparison}: videos corresponding to the ablation study in Fig.~4, illustrating how removing each component degrades trajectory quality in practice.
  \item \textbf{Open-ended prompt results}: full trajectory videos generated from abstract, open-ended prompts, paired with synchronized subtitles and voiceover produced by the Subtitle and Voiceover Generator (\S3.5), demonstrating the complete audiovisual output of our pipeline.
\end{itemize}

\section{Additional Results}
\setcounter{figure}{0}
\setcounter{table}{0}

\subsection{Additional Qualitative Results}

\begin{figure*}[!ht]
  \centering
  \includegraphics[width=0.85\linewidth]{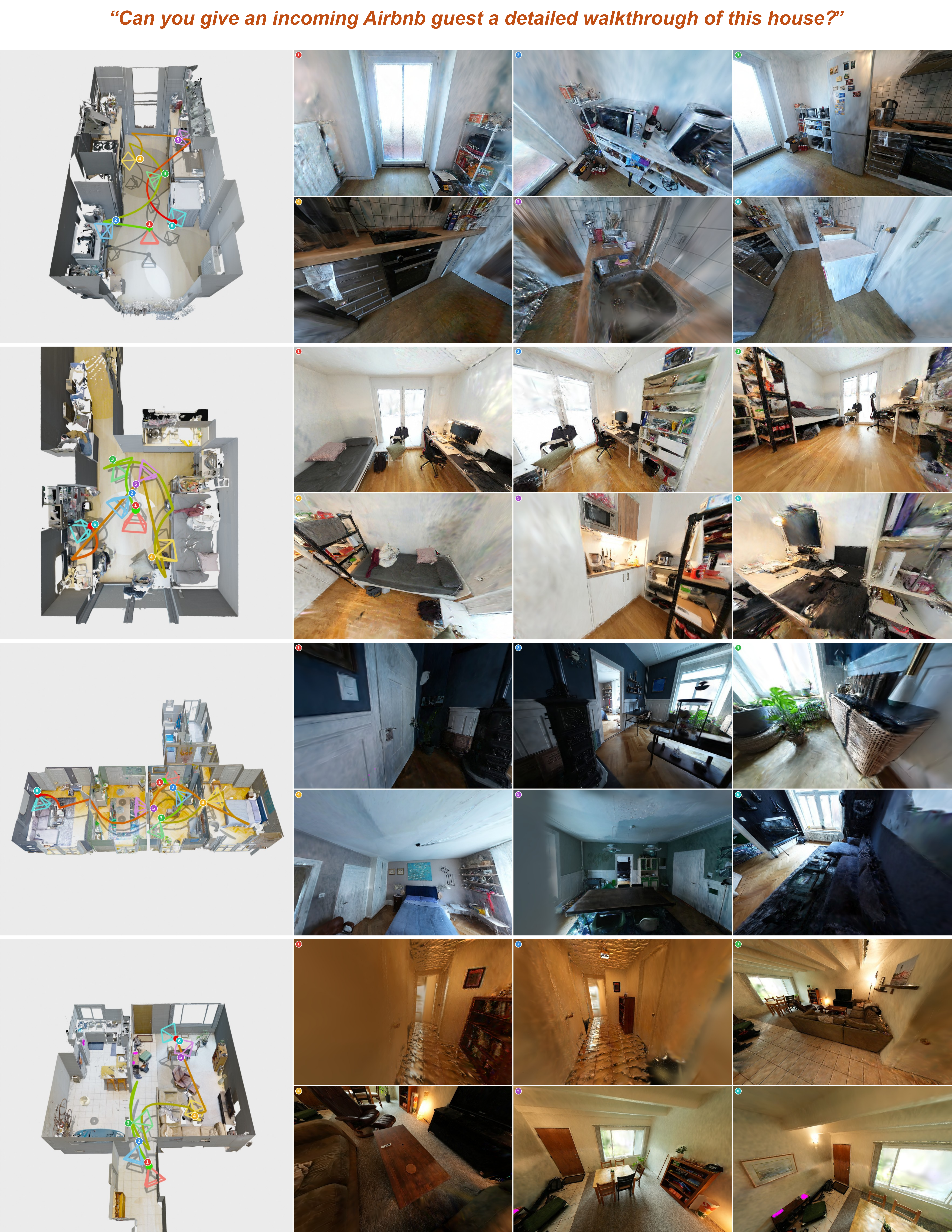}
  \caption{\textbf{Additional results on open-ended prompts.} The User Prompt Translator interprets abstract instructions by leveraging the scene graph to select cinematographically meaningful objects and appropriate camera movements, producing diverse and scene-aware trajectory plans.}
  \Description{Four example scenes answering the prompt: Can you give an incoming Airbnb guest a detailed walkthrough of this house? Each example pairs a top-down view of a reconstructed home, overlaid with a numbered color-coded camera path and camera frustums, with six rendered keyframes taken at the numbered stops. The keyframes show entrances, kitchen counters and appliances, refrigerators, sinks, beds, desks, sofas, and dining areas, illustrating that the planner picks a logical room-by-room visiting order and frames a salient object at each stop.}
  \label{fig:open_ended}
\end{figure*}

Fig.~\ref{fig:open_ended} presents additional trajectory results generated from open-ended prompts on diverse ScanNet++~\cite{yeshwanth2023scannet++} scenes. These results complement the fully-specified comparisons in the main paper by demonstrating the system's autonomous planning capability. Given only abstract instructions (e.g., ``Can you give an incoming Airbnb guest a detailed walkthrough of this house?''), the User Prompt Translator (\S3.2) identifies cinematographically relevant objects from the scene graph, determines an appropriate visitation order, and selects per-object camera movements that match the requested mood --- all without explicit object or movement specifications from the user.

\subsection{Additional Quantitative Results}
\setcounter{figure}{0}
\setcounter{table}{0}

\begin{table}[H]
\centering
\caption{\textbf{Quantitative ablation for partially-specified and open-ended prompts} on ScanNet++~\cite{yeshwanth2023scannet++}.}
\label{tab:medium_high}
\setlength{\tabcolsep}{4pt}
\resizebox{\linewidth}{!}{%
\begin{tabular}{l ccccc}
\toprule
\multirow{2}{*}{\textbf{Method}} & \textbf{CLaTr}~\cite{e.t.} & \textbf{Collision} & \textbf{Occlusion} & \textbf{Coverage} \\
 & \textbf{Score}\,$\uparrow$ & \textbf{Rate}\,$\downarrow$ & \textbf{Rate}\,$\downarrow$ & \textbf{Rate}\,$\uparrow$ \\
\midrule
\multicolumn{5}{l}{\textit{Partially-specified prompts}} \\
\midrule
w/o Anchor Selector                  & 9.055      & 0.026    & 0.658    & 0.857             \\
w/o Parametric Traj.                 & 7.475      & 0.047    & 0.464    & 1.000             \\
w/o SDF-based Opt.                   & 13.501     & 0.512    & 0.597    & 1.000             \\
\textbf{\model{} (Ours)}             & 10.488     & 0.070    & 0.541    & 1.000             \\
\midrule
\multicolumn{5}{l}{\textit{Open-ended prompts}} \\
\midrule
w/o Anchor Selector                  & N/A        & 0.025    & 0.322    & 0.840             \\
w/o Parametric Traj.                 & 10.145     & 0.031    & 0.185    & 1.000             \\
w/o SDF-based Opt.                   & N/A        & 0.459    & 0.366    & 1.000             \\
\textbf{\model{} (Ours)}             & N/A        & 0.040    & 0.199    & 1.000             \\
\bottomrule
\end{tabular}%
}
\end{table}

The main paper evaluates ablations on fully-specified prompts (\S4.2), where both target objects and camera movements are explicitly given. Table~\ref{tab:medium_high} extends this analysis to the two less constrained prompt types --- partially-specified (objects named, movements left to the system) and open-ended (abstract requests with no object or movement specification) --- to verify that the contribution of each component is robust across varying levels of user specificity.

The three ablation trends identified under fully-specified prompts (Table 1) persist consistently across both additional prompt types.
\textbf{(1)~Removing the Anchor Selector} degrades object coverage in all settings (0.882\,$\to$\,0.857\,$\to$\,0.840 for fully-specified, partially-specified, and open-ended, respectively, vs.\ 1.000 for the full pipeline), confirming that CLIP-based anchor selection is unreliable regardless of how prompts are specified.
\textbf{(2)~Removing SDF-based optimization} causes collision rates to spike dramatically (0.557, 0.512, and 0.459), remaining the worst among all variants across every prompt type. This confirms that the 3DGS density field is a poor proxy for solid geometry irrespective of the planning complexity.
\textbf{(3)~Removing parametric trajectories} yields the lowest occlusion rates (0.460, 0.464, 0.185) since unconstrained 6-DoF diffusion poses can freely orient toward targets, at the expense of cinematographic structure.
In all three settings, the full \model{} pipeline achieves the best overall balance: perfect object coverage (1.000), competitive collision and occlusion rates, and coherent cinematographic motion.

Two metric-level differences from the main-paper evaluation are worth noting. First, \emph{Motion MSE is not reported} because partially-specified and open-ended prompts do not prescribe exact camera movements, and thus lack ground-truth trajectories for comparison. Second, \emph{CLaTr scores are marked N/A} for several open-ended configurations. CLaTr~\cite{e.t.} is trained on fully-specified trajectory descriptions (e.g., ``orbit around the table''), and this domain gap causes the text-trajectory cosine similarity to become \emph{negative} when evaluated on abstract, open-ended prompts (e.g., ``Create a cinematic tour of this interior'') whose embeddings fall outside the learned distribution. We therefore report CLaTr only where the similarity is non-negative and semantically interpretable.

\subsection{Runtime Analysis}

\begin{table}[H]
\centering
\caption{\textbf{Runtime comparison of geometry representations} for collision-free trajectory optimization on ScanNet++~\cite{yeshwanth2023scannet++}. Times are averaged over all 50 scenes and include both geometry construction and per-segment optimization.}
\label{tab:sdf_density_runtime}
\setlength{\tabcolsep}{8pt}
\begin{tabular}{l c}
\toprule
{\textbf{Method}} & \textbf{Avg.\ time per scene (s)} \\
\midrule
3DGS Density                 & 450             \\
\textbf{SDF (ours)}          & 10              \\
\bottomrule
\end{tabular}
\end{table}

Table~\ref{tab:sdf_density_runtime} compares the average per-scene runtime of our SDF-based optimizer against an alternative that uses the raw 3DGS~\cite{3dgs} density field as the collision proxy. The SDF representation achieves an $45{\times}$ speedup. This gap arises from two factors: (1)~the SDF is precomputed once as a compact $256^3$ grid and queried via trilinear interpolation, whereas density-based optimization requires evaluating 3DGS splatting at every sample point during each iteration; and (2)~the smooth SDF gradients lead to faster convergence (fewer iterations to clear collisions), while the noisy density gradients often trigger the adaptive weight-doubling mechanism described in \S D.5, prolonging optimization. In practice, the full \model{} pipeline (without scene graph extraction, which takes around 10 minutes per scene) --- including LLM planning, trajectory building, and two-pass optimization --- completes in under 20\,s per scene on a single NIVIDA RTX 5090 GPU. Moreover, due to the sparsity of 3dgs density, sometimes the optimizer will wrongly think that there are not collisions while actually there are.

\section{User Intention Interpretation}
\setcounter{figure}{0}
\setcounter{table}{0}

\begin{figure*}[!ht]
  \centering
  \includegraphics[width=0.85\linewidth]{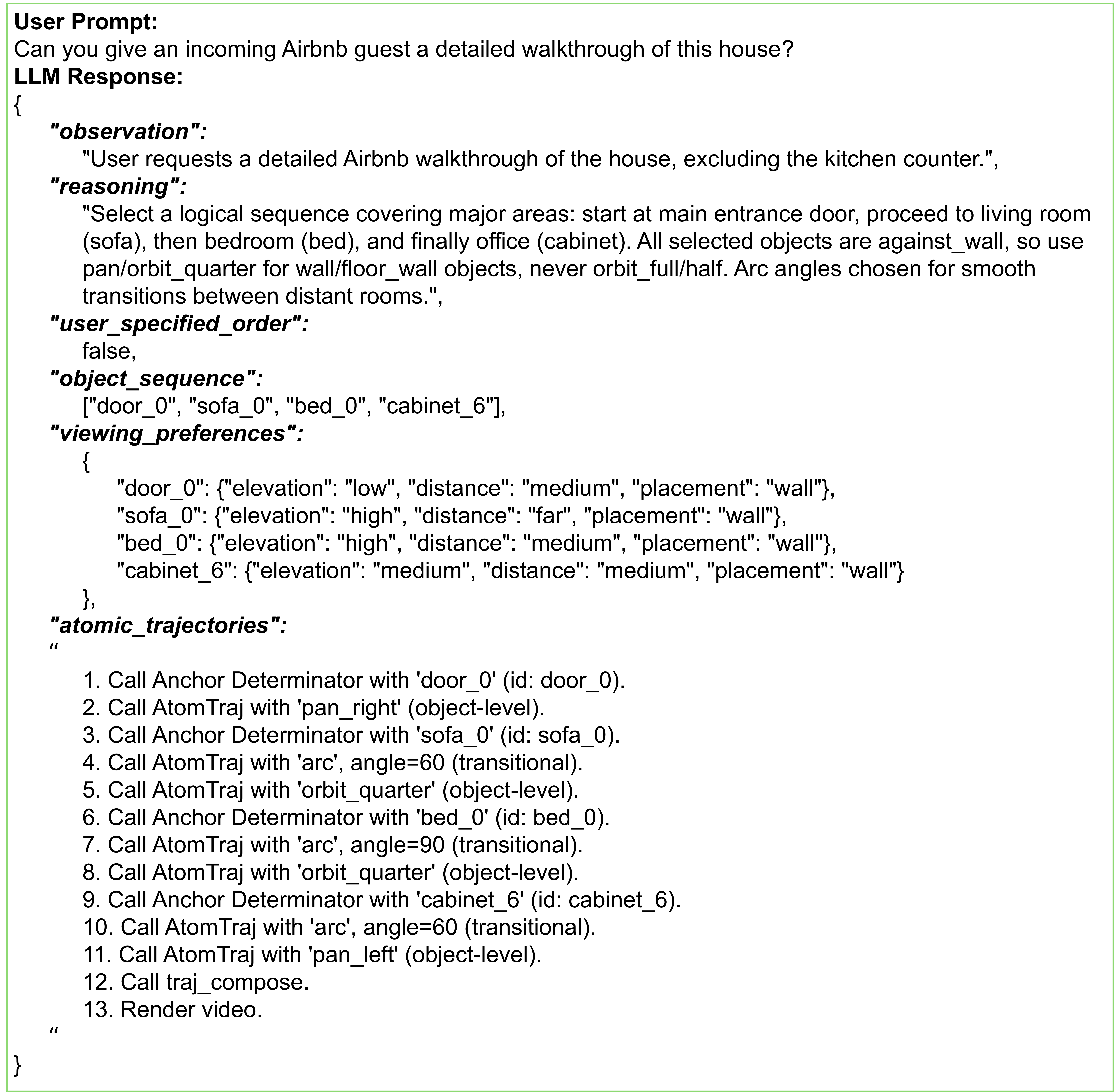}
  \caption{\textbf{User prompt interpretation example.} Given an open-ended prompt and the scene graph, the LLM agent reasons about cinematographically relevant objects, selects appropriate movements respecting placement constraints, and outputs a structured atomic trajectory plan.}
  \Description{A structured LLM response for the prompt: Can you give an incoming Airbnb guest a detailed walkthrough of this house? The JSON-style output contains an observation field; a reasoning field that selects a logical sequence starting at the main entrance door, proceeding to the living room sofa, then the bedroom bed, then the office cabinet, and notes that wall-placed objects only allow pan or quarter-orbit movements; a user-specified-order flag set to false; an object sequence listing door, sofa, bed, and cabinet identifiers; per-object viewing preferences for elevation, distance, and placement; and a thirteen-step atomic trajectory program that alternates anchor determinator calls with atomic trajectory calls such as pan right, arc with a given angle, and quarter orbit, ending with trajectory composition and video rendering.}
  \label{fig:intention_interpretation}
\end{figure*}

A key capability of \model{} is translating abstract, open-ended user prompts into concrete cinematographic plans. Fig.~\ref{fig:intention_interpretation} illustrates the full reasoning chain of the User Prompt Translator on an open-ended input. Given the prompt and the scene graph, the LLM first identifies which objects are cinematographically relevant based on the scene context, then selects an appropriate visitation order and per-object camera movements that match the requested mood. Notably, the agent respects placement constraints throughout: wall-mounted objects are assigned pan movements rather than full orbits, and ceiling-attached objects are excluded from crane shots. This example demonstrates that the structured scene graph and cinematographic toolset enable the LLM to produce coherent, constraint-aware plans even without explicit object or movement specifications from the user.

\section{Additional Implementation Details}
\setcounter{figure}{0}
\setcounter{table}{0}

\subsection{Atomic Trajectory Types}

\begin{figure*}[!ht]
  \centering
  \includegraphics[width=0.68\linewidth]{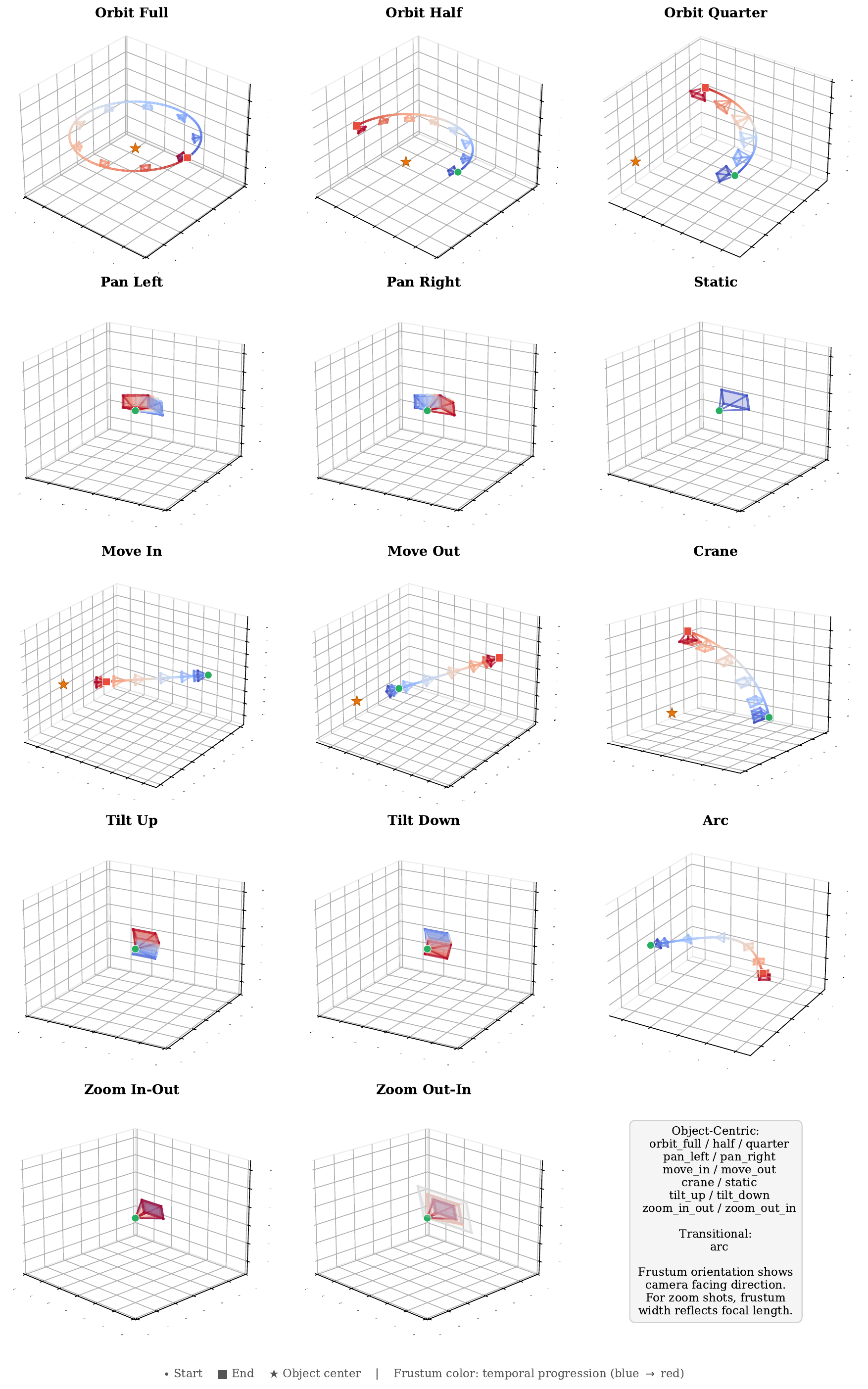}
  \caption{\textbf{Visualization of all 13 camera movement types} in the cinematographic toolset. Each movement is shown with its characteristic trajectory (colored curve) and target object (shaded region).}
  \Description{A grid of small 3D plots, one per movement type, covering orbit full, orbit half, orbit quarter, pan left, pan right, static, move in, move out, crane, tilt up, tilt down, arc, zoom in-out, and zoom out-in. Each plot shows a sequence of camera frustums whose color encodes temporal progression from blue at the start to red at the end, with a dot marking the start pose, a square the end pose, and a star the target object center. A legend panel groups the movements into object-centric types and the transitional arc type, and notes that for zoom shots the frustum width reflects the changing focal length.}
  \label{fig:camera_movements_all}
\end{figure*}

We visualize all camera movement types in our toolset in Fig.~\ref{fig:camera_movements_all}.

\subsection{LLM and Vision-Language Models}
We use GPT-4.1~\cite{gpt4} as the LLM backbone for the User Prompt Translator throughout all experiments. For subtitle and voiceover generation (\S3.5), we employ GPT-4o as the vision-language model to produce per-segment captions from rendered keyframes, and GPT-4o-mini-TTS for text-to-speech synthesis. Keyframes are extracted at 2\,fps from the rendered video and resized to a maximum dimension of 512 pixels. The video is divided into segments of approximately 4 seconds each, with exactly one subtitle and one voiceover line generated per segment to ensure temporal synchronization.

\subsection{Anchor Determination}
For each target object, the Anchor Selector generates candidate viewpoints via face-normal-biased sampling on the six faces of the object's oriented bounding box. For each accessible face, candidates are generated at four distance factors $\{0.5, 0.7, 1.0, 1.3\}$ relative to the preferred viewing distance, ten azimuthal offsets uniformly spaced within $\pm 45^{\circ}$, and three elevation angles derived from the LLM-specified elevation preference. This yields up to 720 candidates per object from face-normal sampling, supplemented by a sparser set of uniform candidates around the object center. Candidates outside the scene bounds (with a 0.3\,m margin) are discarded, and the remaining candidates are filtered for collisions using the precomputed SDF grid with a collision margin of 0.10\,m. The surviving candidates are scored using Equation~2, with the weights specified in \S4.1.1. The viewing distance is estimated from the median bounding-box extent, scaled by a factor of 1.2, and clamped to $[0.3, 3.0]$\,m, with the upper bound further limited to 40\% of the smallest horizontal room dimension to keep cameras well inside the room. The LLM specifies symbolic elevation preferences mapped to angular ranges: \texttt{low}~$(10^{\circ}$--$20^{\circ})$, \texttt{medium}~$(25^{\circ}$--$35^{\circ})$, \texttt{high}~$(40^{\circ}$--$55^{\circ})$, and \texttt{overhead}~$(60^{\circ}$--$80^{\circ})$; distance preferences are mapped to multipliers: \texttt{close}~${\times}\,0.7$, \texttt{medium}~${\times}\,1.0$, \texttt{far}~${\times}\,1.4$. When no preference is specified, the elevation range falls back to a geometry-based heuristic: objects with a horizontal-to-vertical aspect ratio above 2.5 default to \texttt{high}, below 0.6 to \texttt{low}, and otherwise to $(15^{\circ}$--$30^{\circ})$.

\subsection{SDF Construction}
We construct a watertight signed distance field from the scene graph's oriented bounding boxes through a four-stage pipeline: adaptive voxelization, morphological closing, marching cubes surface extraction, and flood-fill signed distance computation. The full pipeline details and a visual illustration are provided in \S\ref{sec:sdf_pipeline}.

\subsection{Trajectory Optimization}
The two-pass optimizer uses Adam with an initial learning rate of 1.0, running for up to 1{,}000 iterations per segment. A ReduceLROnPlateau scheduler halves the learning rate after 30 iterations without improvement. Gradients are clipped to a maximum norm of 1.0. The cost terms and their default weights are: SDF collision cost $\lambda_\text{sdf}=5.0$ with a safety margin $d_\text{safe}=0.2$\,m, occlusion cost $\lambda_\text{occl}=3.0$, smoothness (jerk minimization) $\lambda_\text{smooth}=0.5$, parameter regularization $\lambda_\text{reg}=0.1$ (range-normalized), and boundary cost $\lambda_\text{bnd}=5.0$ (exponential barrier at scene edges). For the occlusion cost, rays are marched in $K=32$ steps from each camera position toward the target center, with a near-skip fraction of 0.1 to avoid self-intersection and an SDF hit threshold $\delta_\text{occ}=0.05$\,m. Target object self-occlusion is prevented by masking ray samples within 50\% of the target OBB's maximum half-extent. The transmittance-weighted accumulation (Equation~7) uses $\text{softplus}$ with $\beta=20$. Optimization terminates early when both collision and occlusion counts reach zero, or when the cost plateaus over 10--20 iterations. If the collision rate remains above 5\% at plateau, $\lambda_\text{sdf}$ is doubled (up to $10^3$) and optimization continues. In Pass~1, object-level segments are optimized with all costs including occlusion; only a subset of free parameters is adjusted per trajectory type (e.g., \texttt{end\_angle} and \texttt{radius} for orbits, \texttt{start\_radius} and \texttt{end\_radius} for dollies) to preserve shot style. In Pass~2, transitional arcs are rebuilt from updated endpoints and optimized over the arc angle parameter only, without occlusion cost. For cross-room transitions, arcs are split into sub-arcs through door/window waypoints from the room connectivity graph, each allocated 30 frames and initialized with zero curvature.

\section{LLM Agent Prompt Design}
\setcounter{figure}{0}
\setcounter{table}{0}

We provide the full system prompt structure used by the User Prompt Translator (\S3.2). The LLM (GPT-4.1, temperature 0.1) receives a system prompt comprising four components: (1) the cinematographic toolset specification, (2) the viewing preference instructions, (3) the 1-3-1 trajectory pattern definition, and (4) eight in-context examples covering diverse scenarios.

\subsection{Cinematographic Toolset}
The system prompt defines the three-layer toolset as follows:

\paragraph{Object-level trajectories $\mathcal{T}_o$.} The LLM selects from 13 movement types: \texttt{orbit\_full}, \texttt{orbit\_half}, \texttt{orbit\_quarter} (circular arcs of $360^{\circ}$, $180^{\circ}$, $90^{\circ}$), \texttt{pan\_left}, \texttt{pan\_right} (stationary yaw rotation), \texttt{move\_in}, \texttt{move\_out} (radial dolly), \texttt{zoom\_in\_out}, \texttt{zoom\_out\_in} (optical focal-length change with automatic return), \texttt{crane} (ascending orbital arc to overhead view), \texttt{tilt\_up}, \texttt{tilt\_down} (stationary pitch rotation), and \texttt{static} (hold pose). Each movement includes a natural-language description of its visual semantics to guide LLM selection.

\paragraph{Transitional trajectories $\mathcal{T}_t$.} A single \texttt{arc} type with a curvature angle parameter $\alpha \in [-89^{\circ}, 89^{\circ}]$, where $\alpha = 0$ produces a straight line and larger magnitudes create sweeping curves. Positive values curve left, negative values curve right.

\paragraph{Anchor selector $\mathcal{A}$.} Invoked before each object-level trajectory to determine the optimal viewpoint. The LLM specifies per-object viewing preferences: elevation $\in$ \{\texttt{low}, \texttt{medium}, \texttt{high}, \texttt{overhead}\}, distance $\in$ \{\texttt{close}, \texttt{medium}, \texttt{far}\}, and placement $\in$ \{\texttt{freestanding}, \texttt{wall}, \texttt{ceiling}, \texttt{floor\_wall}\}.

\subsection{Placement Constraints}
The system prompt encodes hard constraints derived from the scene graph's \texttt{against\_wall} and \texttt{attached\_to\_ceiling} flags:

\begin{itemize}[nosep,leftmargin=1.5em]
  \item \texttt{wall} placement (\texttt{against\_wall=true}): \texttt{orbit\_full/half} are forbidden.
  \item \texttt{ceiling} placement (\texttt{attached\_to\_ceiling=true}): \texttt{crane} is forbidden.
  \item \texttt{floor\_wall} placement (floor-standing, \texttt{against\_wall=true}): \texttt{orbit\_full} is forbidden.
  \item \texttt{freestanding} placement: all movements are allowed.
\end{itemize}

These rules are reinforced through both explicit instructions (Rules 18--20 in the system prompt) and through all eight in-context examples, which consistently demonstrate correct placement-aware movement selection.

\subsection{Trajectory Pattern}
The prompt enforces the 1-3-1 sequential pattern described in \S3.2.2. For a sequence of $N$ objects, the LLM produces exactly $3N - 1$ atomic steps: one initial anchor call, followed by repeated (object-level $\to$ anchor $\to$ transitional) loops, ending with a final object-level trajectory. The LLM also determines the visitation order: when the user explicitly specifies a sequence, the \texttt{user\_specified\_order} flag is set to \texttt{true}; for open-ended requests, the LLM selects a logical order and sets the flag to \texttt{false}, enabling the downstream nearest-neighbor reordering heuristic (\S3.3.3).

\subsection{In-Context Examples}
The system prompt includes eight examples spanning: multi-object room tours (4 and 6 objects), explicit user-specified paths, single-object inspection, static holds, close-up inspection with \texttt{move\_in}, dramatic reveals with \texttt{move\_out}, and placement-constrained scenarios (wall-mounted objects using \texttt{pan} instead of \texttt{orbit}). Each example demonstrates the complete JSON output format including observation, reasoning, object sequence, viewing preferences with placement labels, and the numbered atomic trajectory steps.

\subsection{User Prompt Construction}
At inference time, the user prompt includes: the natural language request, a JSON list of all scene objects (with ID, label, center position, size, \texttt{against\_wall}, \texttt{attached\_to\_ceiling}, and room assignment), and the room-to-object mapping. The LLM responds with a structured JSON containing the object sequence, per-object viewing preferences, and the numbered atomic trajectory steps, which are then validated for 1-3-1 pattern compliance and placement constraint adherence before execution.

\section{Cross-Room Trajectory Routing}
\setcounter{figure}{0}
\setcounter{table}{0}

When consecutive target objects in a trajectory plan reside in different rooms, a direct transitional arc between them will typically collide with walls. To handle this, we build a room connectivity graph from the scene graph and route transitional arcs through door or doorway waypoints.

\subsection{Room Connectivity Graph}
From the scene graph's room-to-object mapping and the set of passage objects (doors, door frames, archways, openings), we construct an undirected graph where nodes are rooms and edges represent passages connecting them. For each passage object, we determine which two rooms it connects: if the passage is listed in a room, that room is one endpoint, and the nearest other room (by object-center distance) is the second endpoint. For passages not assigned to any room, the two nearest rooms are used. This produces a graph where, for example, a hallway connecting a living room and a kitchen yields edges \texttt{living\_room} $\leftrightarrow$ \texttt{hallway} and \texttt{hallway} $\leftrightarrow$ \texttt{kitchen}, each labeled with the specific door or door frame object at that boundary.

\subsection{Waypoint Selection}
Given two objects in different rooms, we run BFS on the room graph to find the shortest room-hop path, then select the best passage at each hop. When multiple passages connect the same room pair (e.g., a door and a door frame side by side), we score each by a combination of path length and an openness preference: door frames and archways (always open, preference score 0) are strongly preferred over doors (which may be closed, preference score 2). Specifically, the score for passage $p$ at hop $i$ is:
\begin{equation*}
  \text{score}(p) = \|{\mathbf{x}}_\text{curr} - \mathbf{c}_p\| + \|\mathbf{c}_p - {\mathbf{x}}_\text{dest}\| + 0.5 \cdot \text{pref}(p),
\end{equation*}
where $\mathbf{c}_p$ is the passage center, ${\mathbf{x}}_\text{curr}$ is the current camera position, ${\mathbf{x}}_\text{dest}$ is the final destination, and $\text{pref}(p) \in \{0, 1, 2, 5\}$ encodes openness preference. The waypoint height is adjusted to match the average camera height of the neighboring trajectory endpoints, clamped within the door's vertical extent.

\subsection{Arc Splitting}
Once waypoints are determined, the original single transitional arc is replaced by a sequence of sub-arcs: one per hop. Each sub-arc connects consecutive waypoints (or the trajectory endpoint to the first/last waypoint), is allocated 30 frames, and is initialized with zero curvature ($\alpha = 0$, i.e., a straight line through the doorway). These sub-arcs are then individually optimized in Pass~2 of the trajectory optimizer over the arc angle parameter only, ensuring collision-free passage through each doorway while allowing gentle curves where space permits. This approach naturally handles both simple two-room transitions (one door, one sub-arc) and multi-room routes through hallways or corridors (multiple doors, multiple sub-arcs).

\section{Benchmark Prompt Generation}
\setcounter{figure}{0}
\setcounter{table}{0}

As described in \S4.1.2, we generate benchmark prompts via template-based sampling with a fixed seed for reproducibility. For each of the 50 ScanNet++~\cite{yeshwanth2023scannet++} scenes, we produce two prompts at each of three specificity levels (fully-specified, partially-specified, and open-ended), yielding six prompts per scene (300 total). Objects are selected by first filtering out structural elements (walls, floors, ceilings) and small/uninteresting items (switches, outlets, wires), then sampling 3--6 objects per prompt with label deduplication (at most one object per semantic class) to ensure diversity. For fully-specified prompts, ceiling-attached objects are additionally excluded due to poor 3DGS reconstruction quality near ceilings.

\subsection{Open-Ended Prompts}
Open-ended prompts are sampled from a pool of 20 abstract templates that describe a mood, purpose, or scenario without mentioning any specific object names or camera movement terms. Examples include:

\begin{itemize}[nosep,leftmargin=1.5em]
  \item ``Create a cinematic tour of this interior.''
  \item ``Imagine I'm an Airbnb guest arriving for the first time --- give me a welcome tour.''
  \item ``Film this room like a short architectural showcase.''
\end{itemize}

\subsection{Partially-Specified Prompts}
Partially-specified prompts name $\geq$3 objects by their semantic label but leave camera movement selection to the system. Templates use natural phrasing with placeholders for the object list, first, middle, and last objects. Examples:

\begin{itemize}[nosep,leftmargin=1.5em]
  \item ``Highlight the sofa, dining table, plant, and bookshelf in a smooth sequence.''
  \item ``Take me on a tour starting from the desk, passing by the monitor, and ending at the bed.''
\end{itemize}

\subsection{Fully-Specified Prompts}
Fully-specified prompts name $\geq$3 objects with explicit camera movements for each, respecting placement constraints derived from the scene graph. Each object's \texttt{against\_wall} and \texttt{attached\_to\_ceiling} flags determine which movements are forbidden: wall-adjacent objects forbid \texttt{orbit\_full} (the camera cannot pass behind the wall), and ceiling-attached objects forbid \texttt{crane} (the camera cannot rise above the ceiling). To maximize diversity within each prompt, movements are drawn from distinct type groups (orbit, pan, dolly, tilt, zoom, crane, static), so no two objects in the same prompt receive the same movement type. Steps are joined with transitional language (``First, ..., then ..., and finally ...''). Examples:

\begin{itemize}[nosep,leftmargin=1.5em]
  \item ``Here's what I want: First, do a full 360$^{\circ}$ orbit around the table, then crane up above the sofa, then zoom in on the bicycle, then pan right at the refrigerator, and finally hold a static shot of the sink.''
  \item ``I'd like a specific sequence: First, move in closer to the plant, then do a half orbit around the desk, and finally tilt down toward the chair.''
\end{itemize}

\section{Ground Truth Creation}
\begin{figure}[H]
  \centering
  \includegraphics[width=\linewidth]{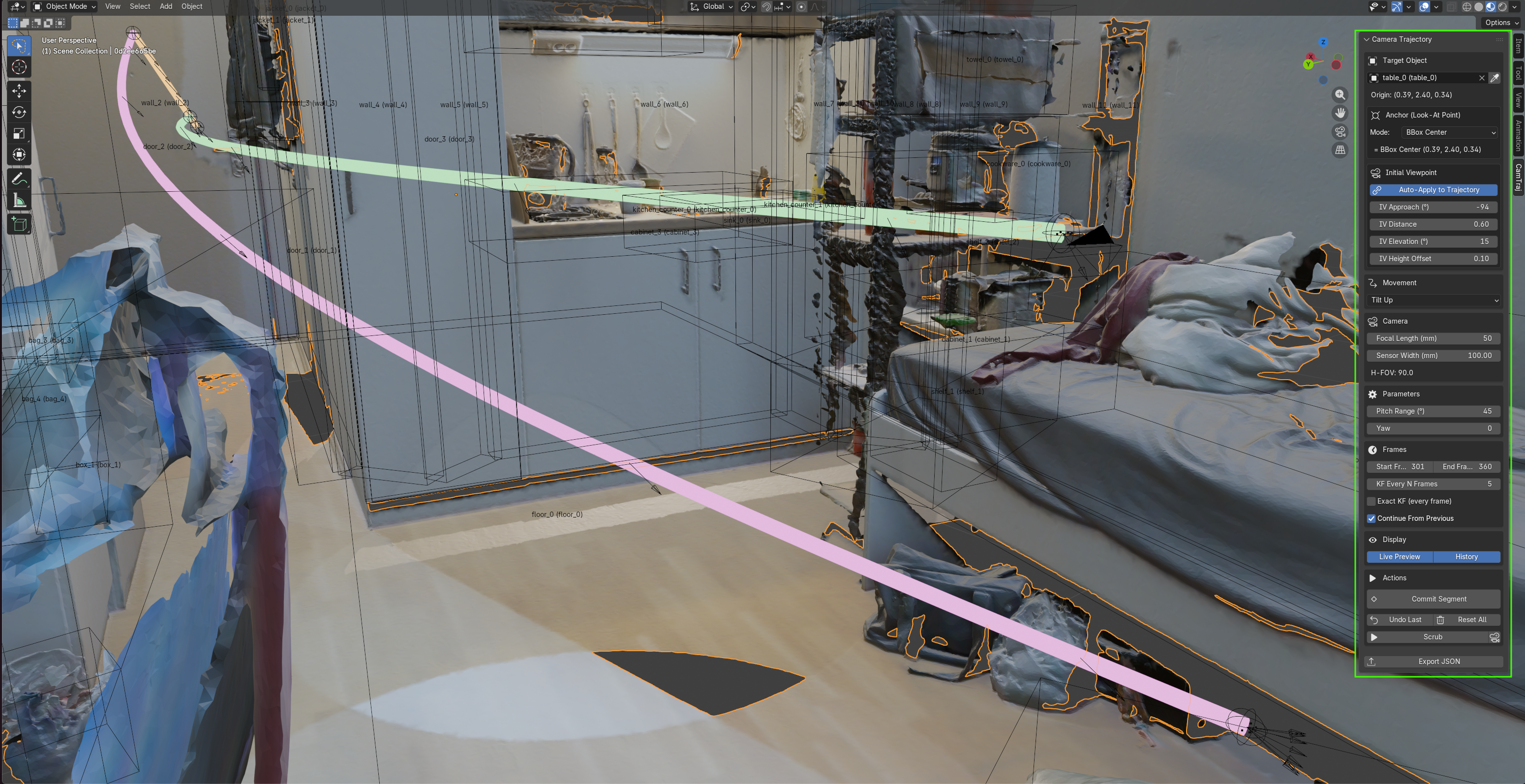}
  \caption{\textbf{Blender addon for ground truth trajectory creation.} The operator selects target objects and camera movement types from the addon panel (green box). The tool generates smooth, collision-free keyframed trajectories that serve as references for the Motion MSE metric.}
  \Description{Screenshot of Blender during trajectory authoring. The 3D viewport shows a reconstructed studio apartment with labeled wireframe bounding boxes around walls, doors, cabinets, and furniture; two committed trajectory segments are drawn as thick curves, one green and one pink, each ending at a small camera frustum. The addon panel on the right, highlighted with a green border, exposes controls for target object selection, look-at anchor mode, initial viewpoint approach angle, distance, elevation and height offset, movement type with tilt up currently selected, camera focal length and sensor width, pitch and yaw parameters, frame range and keyframe frequency, live preview and history display, and buttons to commit a segment, undo, reset, scrub, and export the trajectory as JSON.}
  \label{fig:gt_creation}
\end{figure}

Evaluating camera trajectory quality requires ground-truth references, yet no existing dataset provides object-centric cinematographic trajectories in real 3D scenes. To address this, we developed a custom Blender addon (Fig.~\ref{fig:gt_creation}) that streamlines manual trajectory creation. The operator loads the reconstructed 3D scene, selects target objects from a list, and specifies the desired camera movement type for each segment (e.g., orbit, dolly, crane). The addon then generates smooth keyframed trajectories that conform to the selected movement profiles while keeping the target object centered in the frame, then the parameters of the generated trajectories can be modified manually to achieve collision-free effects. For each of the 50 ScanNet++~\cite{yeshwanth2023scannet++} scenes, two fully-specified prompts are created, and the corresponding ground-truth trajectories are generated using this tool by an experienced operator. The resulting trajectories serve as references for the Motion MSE metric~\cite{chatcam} reported in Table~1.

\section{User Study Details}
\setcounter{figure}{0}
\setcounter{table}{0}

\begin{figure}[H]
  \centering
  \includegraphics[width=\linewidth]{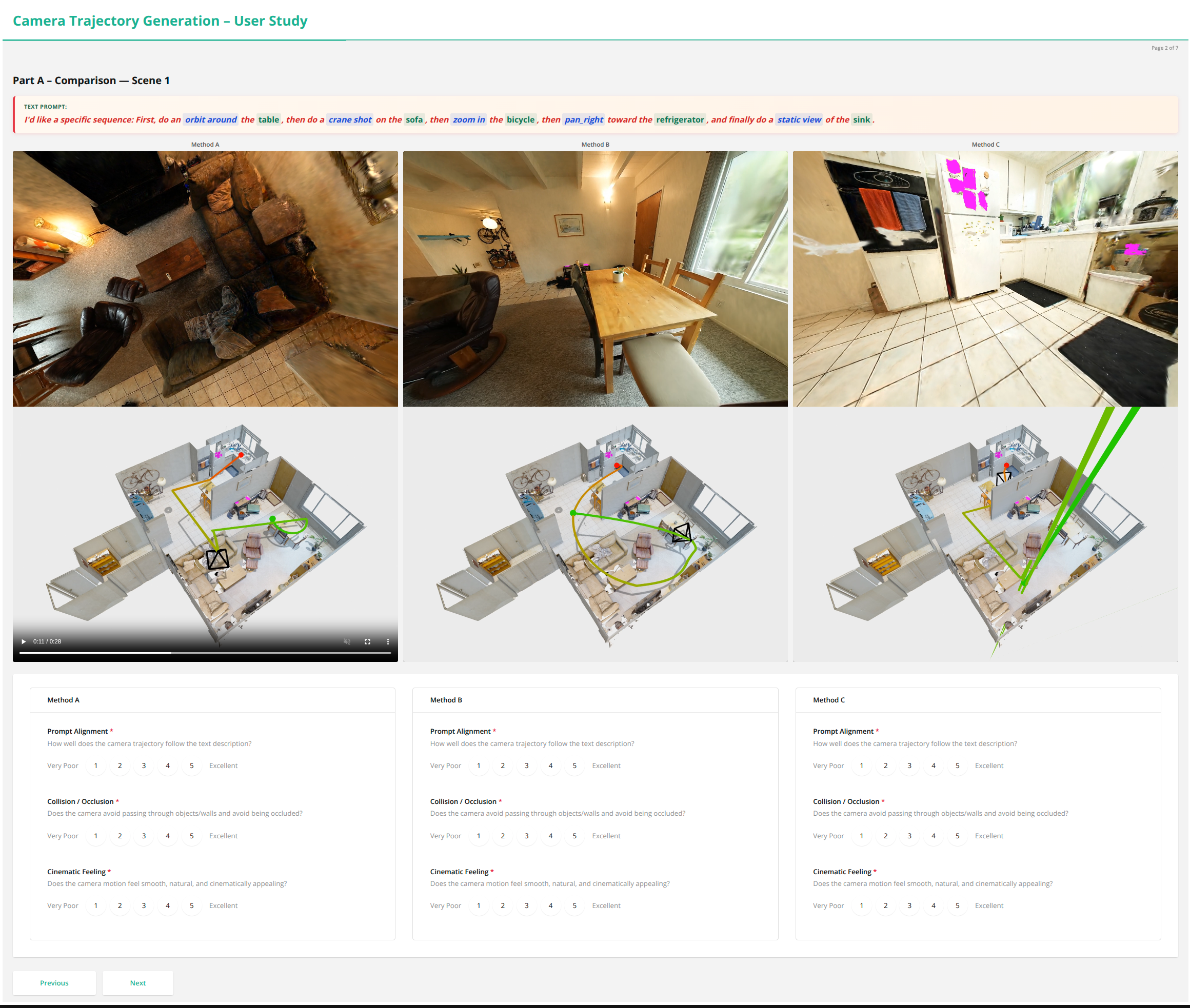}
  \caption{\textbf{User study interface.} Participants view trajectory-rendered videos and rate each on three criteria (Prompt Alignment, Collision \& Occlusion Avoidance, Cinematic Quality) using a 5-point Likert scale. Videos are presented in randomized order within each scene.}
  \Description{Screenshot of one page of the web-based study interface titled Camera Trajectory Generation User Study. Below the text prompt, three columns labeled Method A, Method B, and Method C each contain a rendered first-person video and a top-down trajectory visualization of the same apartment scene. Under each column is a rating form with three required five-point scales ranging from very poor to excellent, covering prompt alignment, collision and occlusion, and cinematic feeling. Previous and next navigation buttons appear at the bottom of the page, and method identities are anonymized.}
  \label{fig:user_study}
\end{figure}

We conduct a perceptual user study with 40 participants. Each participant views trajectory-rendered videos from all six methods (two baselines and four variants including ours) across three representative ScanNet++~\cite{yeshwanth2023scannet++} scenes, yielding 18 ratings per participant. Videos within each scene are presented in randomized order to mitigate position bias. As shown in Fig.~\ref{fig:user_study}, participants rate each video on three criteria using a 5-point Likert scale:
\begin{itemize}[nosep,leftmargin=1.5em]
  \item \textbf{Prompt Alignment} (1--5): how well the camera trajectory follows the requested movements and targets the specified objects.
  \item \textbf{Collision \& Occlusion Avoidance} (1--5): whether the camera avoids passing through scene geometry and maintains clear views of target objects.
  \item \textbf{Cinematic Quality} (1--5): overall smoothness, visual appeal, and professional quality of the camera work.
\end{itemize}
The study is conducted via a web interface on standard desktop monitors. No time limit is imposed per video, and participants may replay videos before rating. Results are reported in Table~2 of the main paper.

\section{Watertight SDF Construction}
\label{sec:sdf_pipeline}
\setcounter{figure}{0}
\setcounter{table}{0}

\begin{figure}[H]
  \centering
  \includegraphics[width=0.75\linewidth]{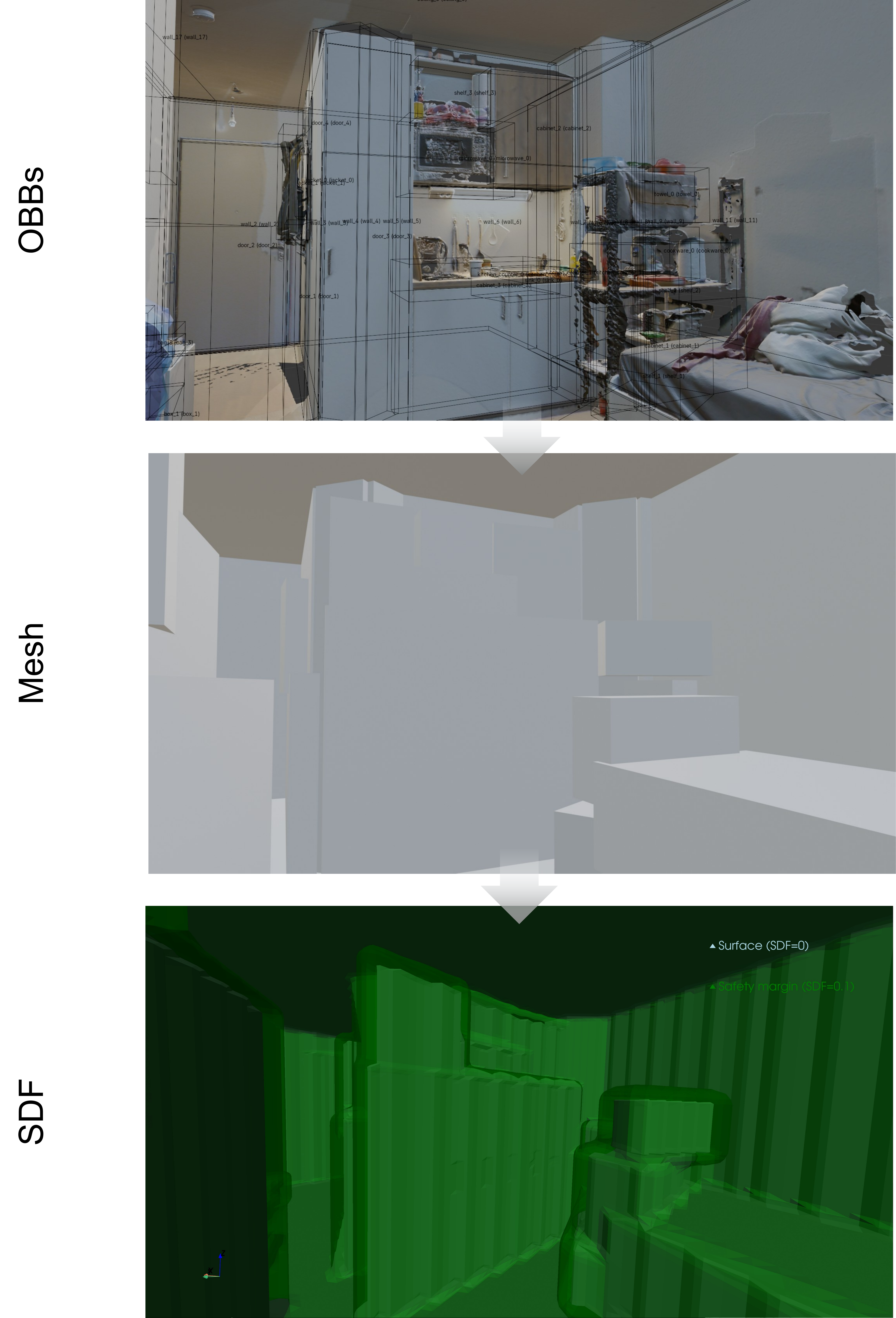}
  \caption{\textbf{Watertight SDF construction pipeline.} Starting from the scene graph's oriented bounding boxes, we voxelize each OBB, apply morphological closing to seal thin-wall gaps, extract a watertight surface via marching cubes, and compute a signed distance field with flood-fill interior detection. The resulting SDF provides smooth, differentiable collision gradients for trajectory optimization.}
  \Description{Three vertically stacked renderings of the same kitchen corner illustrate successive geometry conversion stages. Top, labeled OBBs: the textured scan overlaid with labeled wireframe oriented bounding boxes for walls, doors, cabinets, shelves, and appliances. Middle, labeled Mesh: the boxes fused into a plain gray watertight surface. Bottom, labeled SDF: a translucent green visualization of the signed distance field around the geometry, with annotations marking the zero-level surface and the safety margin level set at SDF equal to 0.1 that is used by the collision cost.}
  \label{fig:sdf_construction}
\end{figure}

The trajectory optimizer requires a signed distance field (SDF) that reliably distinguishes free space (positive SDF) from occupied regions (negative SDF). This in turn requires a \emph{watertight} mesh: any gap in the surface allows the flood-fill interior detection to leak, incorrectly labeling obstacle voxels as free space and permitting collisions.

Fig.~\ref{fig:sdf_construction} illustrates the three-stage pipeline. First, the scene graph extractor ensures structural completeness by extracting wall, floor, and ceiling surfaces and snapping adjacent structural boxes to eliminate boundary gaps (\S3.1). Each oriented bounding box is then voxelized with an adaptive resolution (at least 3 voxels along the thinnest dimension, maximum grid size $512^3$), supplemented by surface sampling to capture thin features such as doors and window frames. One iteration of morphological closing seals remaining thin-wall gaps. Marching cubes extracts a watertight surface from the occupancy field at a level of 0.5, followed by one iteration of Laplacian smoothing to reduce staircase artifacts. Finally, the signed distance field is computed at $256^3$ resolution with 0.5\,m padding using MeshLib's unsigned distance computation, with sign determined by flood-fill from the mesh centroid using 6-connectivity.

This OBB-based construction is substantially faster than computing an SDF from the raw 3DGS point cloud (25\,s vs.\ 450\,s per scene, Table~\ref{tab:sdf_density_runtime}), while also avoiding the noisy density artifacts inherent to 3DGS~\cite{3dgs} representations. During optimization, SDF values at arbitrary positions are obtained via differentiable trilinear interpolation on the cached grid, enabling gradient flow to trajectory parameters.

\section{Limitations}
\setcounter{figure}{0}
\setcounter{table}{0}

We identify four main limitations of the current system.

\subsection{Scene Graph Quality}
The quality of the scene graph directly affects both the LLM reasoning process and the final trajectory. Incorrect semantic labels cause the system to target wrong objects (e.g., labeling a ``monitor'' as a ``TV'' leads the LLM to plan shots for a nonexistent television). Inaccurate oriented bounding boxes result in camera trajectories that focus on non-centric viewpoints or miss the object entirely. When a user-specified object is absent from the scene graph, the LLM falls back to selecting the most semantically similar substitute, which may not match the user's intent. Improving the upstream 3D scene understanding pipeline would directly benefit trajectory quality.

\subsection{Static Scene Assumption}
Our method assumes a static scene and does not account for dynamic elements such as moving people, pets, or articulated objects. In practice, indoor scenes often contain such elements, and a production-quality system would need to reason about temporal occlusion and dynamic obstacle avoidance. Extending the SDF-based optimizer to incorporate predicted object motion is a promising direction for future work.

\subsection{Fixed Cinematographic Toolset}
While the parametric trajectory vocabulary covers the most common cinematographic movements, it does not include more advanced techniques such as Steadicam-style floating motion, handheld shake, or rack focus effects. The toolset is extensible by design --- new movement types can be added by defining their parametric representation and registering them with the LLM prompt --- but the current implementation does not cover the full range of professional cinematography.

\subsection{LLM Dependence and Reproducibility}
The User Prompt Translator relies on a commercial LLM (GPT-4.1~\cite{gpt4}), which introduces two concerns: (1)~API costs scale with the number of scenes and prompt complexity, and (2)~model updates or version changes may alter the LLM's output, affecting reproducibility. We mitigate the latter by using a low temperature (0.1) and structured output validation, but fully deterministic behavior cannot be guaranteed across API versions.

\fi

\end{document}